\pdfoutput=1

\documentclass[11pt]{article}

\usepackage[preprint]{coling}

\usepackage{times}
\usepackage{latexsym}
\usepackage{amsmath}
\usepackage{algorithm}
\usepackage{algpseudocode}
\usepackage{dingbat}

\usepackage{graphicx}
\usepackage{caption}
\usepackage{subcaption}
\usepackage{float}
\usepackage{array}
\usepackage{wasysym}

\usepackage{enumitem}

\newcommand\singlestep{\textcolor{black}{\textsc{Single-Step}}}

\newcommand\singlestepvip{\textcolor{black}{\textsc{Single-Step-ViP}}}

\newcommand\multistep{\textcolor{black}{\textsc{Multi-Step}}}

\usepackage[T1]{fontenc}

\usepackage[utf8]{inputenc}

\usepackage{microtype}

\usepackage{inconsolata}

\usepackage{graphicx}

\usepackage{lipsum,lineno}
\usepackage{pifont}
\long\def\loremlines#1{%
    \setbox\one=\vbox {%
      \lipsum%
     }
\setbox\two=\vsplit\one to #1\baselineskip
\unvbox\two}

\usepackage{booktabs}
\definecolor{ForestGreen}{RGB}{34,139,34}
\definecolor{Blue}{RGB}{53, 130, 230}
\definecolor{Yellow}{RGB}{230,212,45}

\title{Towards Efficient and Robust VQA-NLE Data Generation with Large Vision-Language Models}

\author{Patrick Amadeus Irawan$^{1}$, Genta Indra Winata$^{2}$\thanks{\text{ }The work was conducted outside Capital One.}, Samuel Cahyawijaya$^{3}$, \\
\textbf{Ayu Purwarianti}$^{1}$ \\
  $^1$Institut Teknologi Bandung$\quad$$^2$Capital One\\
  $^3$The Hong Kong University of Science and Technology \\
  \texttt{patrickai0309@gmail.com, genta.winata@capitalone.com,} \\ \texttt{scahyawijaya@connect.ust.hk, ayu@informatika.org}}

\begin{document}
\maketitle
\begin{abstract}
Natural Language Explanation (NLE) aims to elucidate the decision-making process by providing detailed, human-friendly explanations in natural language. It helps demystify the decision-making processes of large vision-language models (LVLMs) through the use of language models. While existing methods for creating a Vision Question-Answering with Natural Language Explanation (VQA-NLE) datasets can provide explanations, they heavily rely on human annotations that are time-consuming and costly. In this study, we propose a novel approach that leverages LVLMs to efficiently generate high-quality synthetic VQA-NLE datasets. By evaluating our synthetic data, we showcase how advanced prompting techniques can lead to the production of high-quality VQA-NLE data. Our findings indicate that this proposed method achieves up to 20$\times$ faster than human annotation, with only a minimal decrease in qualitative metrics, achieving robust quality that is nearly equivalent to human-annotated data. Furthermore, we show that incorporating visual prompts significantly enhances the relevance of text generation. Our study paves the way for a more efficient and robust automated generation of multi-modal NLE data, offering a promising solution to the problem.
\end{abstract}

\section{Introduction}

Natural Language Explanation (NLE) is a valuable tool for elucidating a model's decision-making process, thereby enhancing transparency and fostering trust and accountability. This concept has been applied across various machine learning tasks \cite{hendricks2016generatingvisualexplanations, ling-etal-2017-program, kotonya2020explainableautomatedfactcheckingsurvey, aggarwal-etal-2021-explanations, lu2022learnexplainmultimodalreasoning, yang-etal-2015-wikiqa}, including practical applications such as automated driving \cite{kim2018textualexplanationsselfdrivingvehicles} and medical imaging \cite{kayser2022explainingchestxraypathologies}. In the realm of vision-language tasks, explanation-rich datasets like VQA-X~\cite{park2018multimodalexplanationsjustifyingdecisions}, VQA-E~\cite{li2018vqaeexplainingelaboratingenhancing}, VCR~\cite{zellers2019recognition}, e-SNLI-VE~\cite{kayser2021vil}, and GQA-REX~\cite{chen2022rexreasoningawaregroundedexplanation} have been instrumental in advancing vision-language NLE research. These datasets enable a deeper understanding and improved explainability of interactions within the vision-language modality, thereby enhancing the overall effectiveness of NLE in vision-language tasks, especially in Visual Question Answering (VQA).




\begin{figure}[!t]
    \centering
    \includegraphics[width=0.99\linewidth]{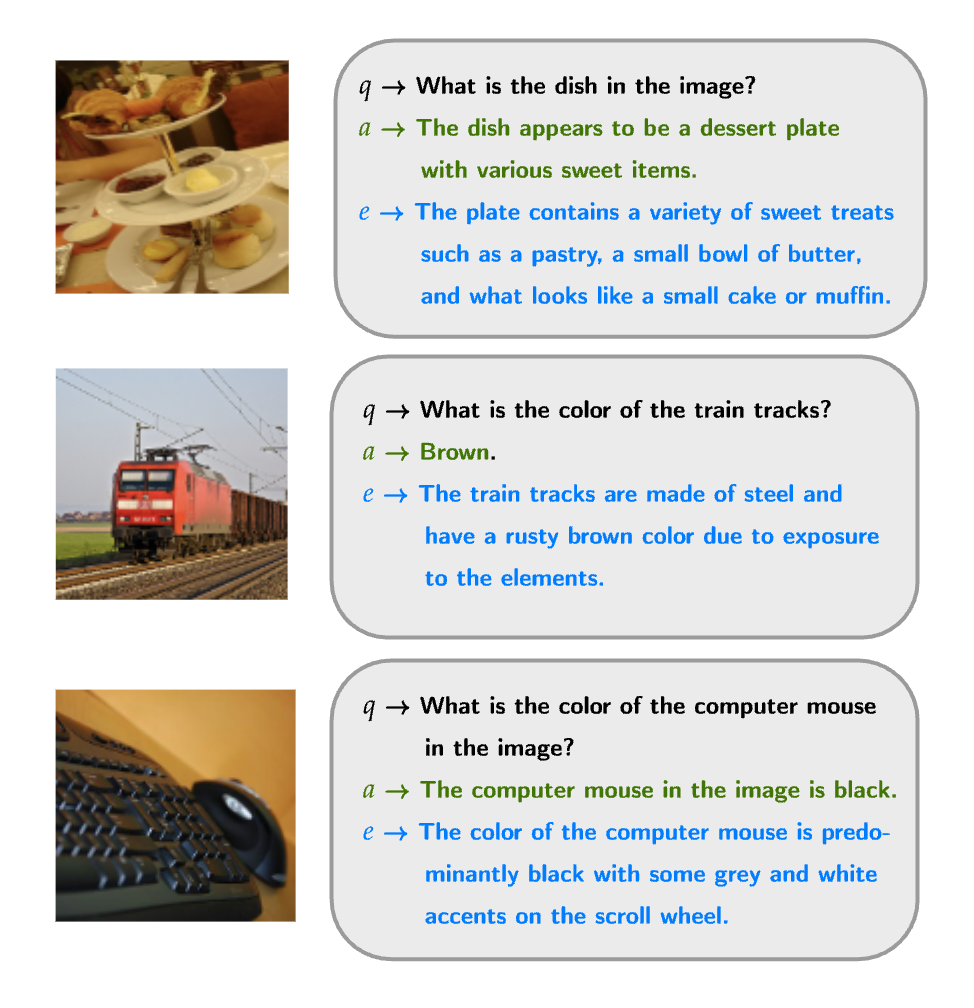} 
    \caption{Generated VQA data along with NLE of the predicted answers, offering better explainability over traditional VQA data. These are the three samples from our synthetic VQA-NLE dataset. We create a total of 66,682 unique instances of these triplets.}
    \label{fig:triplet-example}
\vspace{-5mm}
    
\end{figure}
Despite significant advancements on the topic, the scarcity of VQA-NLE data still prevails, potentially hindering further progress in the field. Existing VQA-NLE datasets~\cite{do2021esnlivecorrectedvisualtextualentailment, park2018multimodalexplanationsjustifyingdecisions, zellers2019recognitioncognitionvisualcommonsense} heavily rely on manual human annotations which is time-consuming and costly. This causes the data creation process inefficiency and difficult to scale underscoring the need for a more efficient method for generate VQA-NLE data~\cite{lu2024machinelearningsyntheticdata,li2018vqaeexplainingelaboratingenhancing,chen2022rexreasoningawaregroundedexplanation}.

In this work, we propose efficient and scalable methods for generating synthetic VQA-NLE data that eliminates the need for additional resource curation while maintaining quality comparable to human-generated data.\footnote{The code is available at \url{https://github.com/patrickamadeus/vqa-nle-llava}. The dataset can be accessed at \url{https://huggingface.co/datasets/patrickamadeus/vqa-nle-llava}} We introduce both single-step and multi-step approaches to produce high-quality data, utilizing visual prompts with bounding boxes to enhance focus and improve generation accuracy. With the advent of large vision language models (LVLMs)~\cite{liu2024improvedbaselinesvisualinstruction, zhu2023minigpt4enhancingvisionlanguageunderstanding,bai2023qwenvlversatilevisionlanguagemodel}, we leverage the generative capabilities of LVLMs to address current limitations for generating synthetic VQA-NLE data. Figure~\ref{fig:triplet-example} showcases the samples of our generated VQA-NLE data. To quantitatively evaluate our method, we create an evaluation dataset and conducted a comparative analysis of various settings. Furthermore, we perform an efficiency analysis against crowdsourced data creation method to reinforce our primary objective of presenting a more efficient method for generating synthetic VQA-NLE datasets. Our contributions are three-fold:
\begin{itemize}
    \item We propose methods to synthetically generate high-quality VQA-NLE data using LVLMs, which show a high correlation with human annotations.
    \item We demonstrate the impact of various synthetic VQA-NLE generation methods to identify best practices for constructing effective and efficient synthetic VQA-NLE data.
    \item We compare the effectiveness and efficiency of our data generation methods against human annotations for the same task. Our findings highlight the strong potential of LVLM-based synthetic VQA-NLE data generation as a viable alternative, producing high-quality data with up to 20$\times$ greater efficiency.
\end{itemize}

\section{Methodology}

\begin{figure*}[!th]
    \centering
    \begin{subfigure}[t]{0.49\textwidth}
        \centering
        \caption{\textsc{Single-Step}}
        \includegraphics[width=\linewidth]{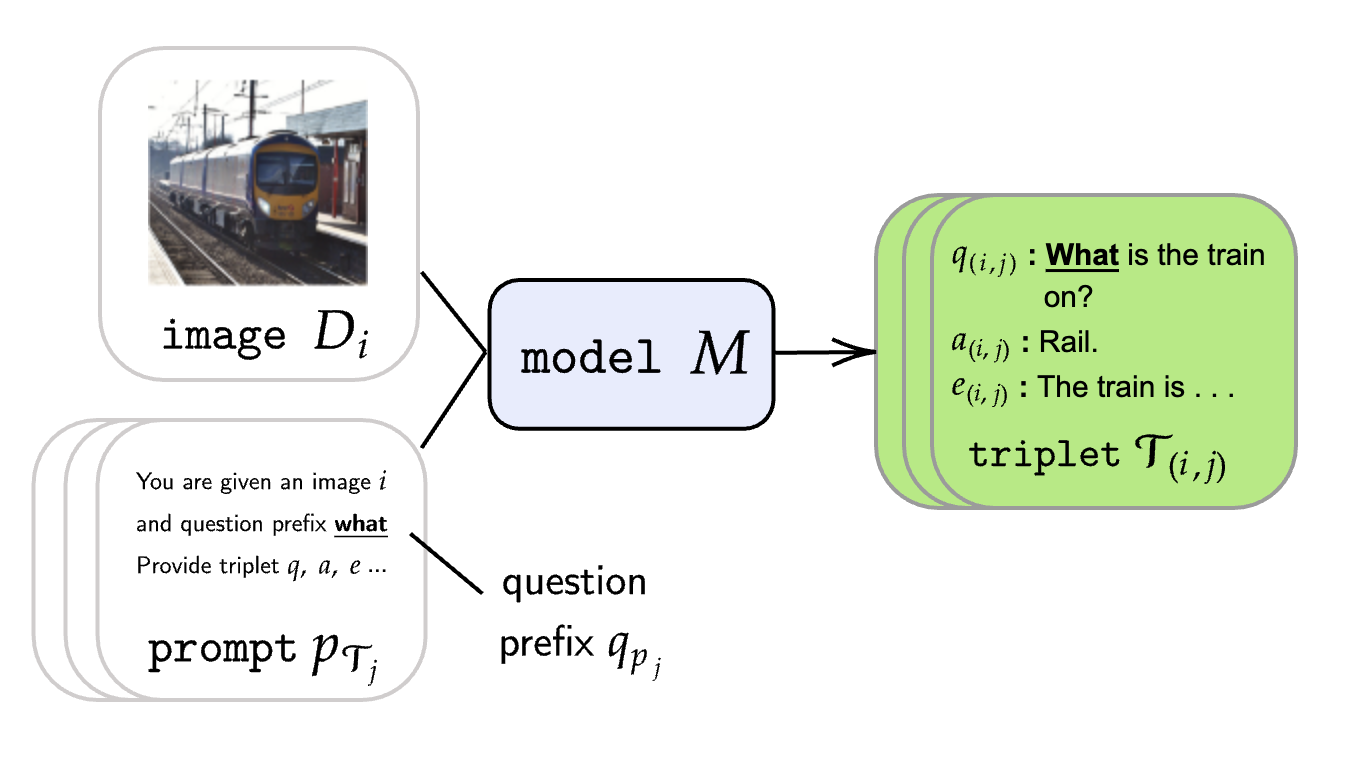} 
    \end{subfigure}
    \begin{subfigure}[t]{0.49\textwidth}
        \centering
        \caption{$\textsc{Single-Step-ViP}$}
        \includegraphics[width=\linewidth]{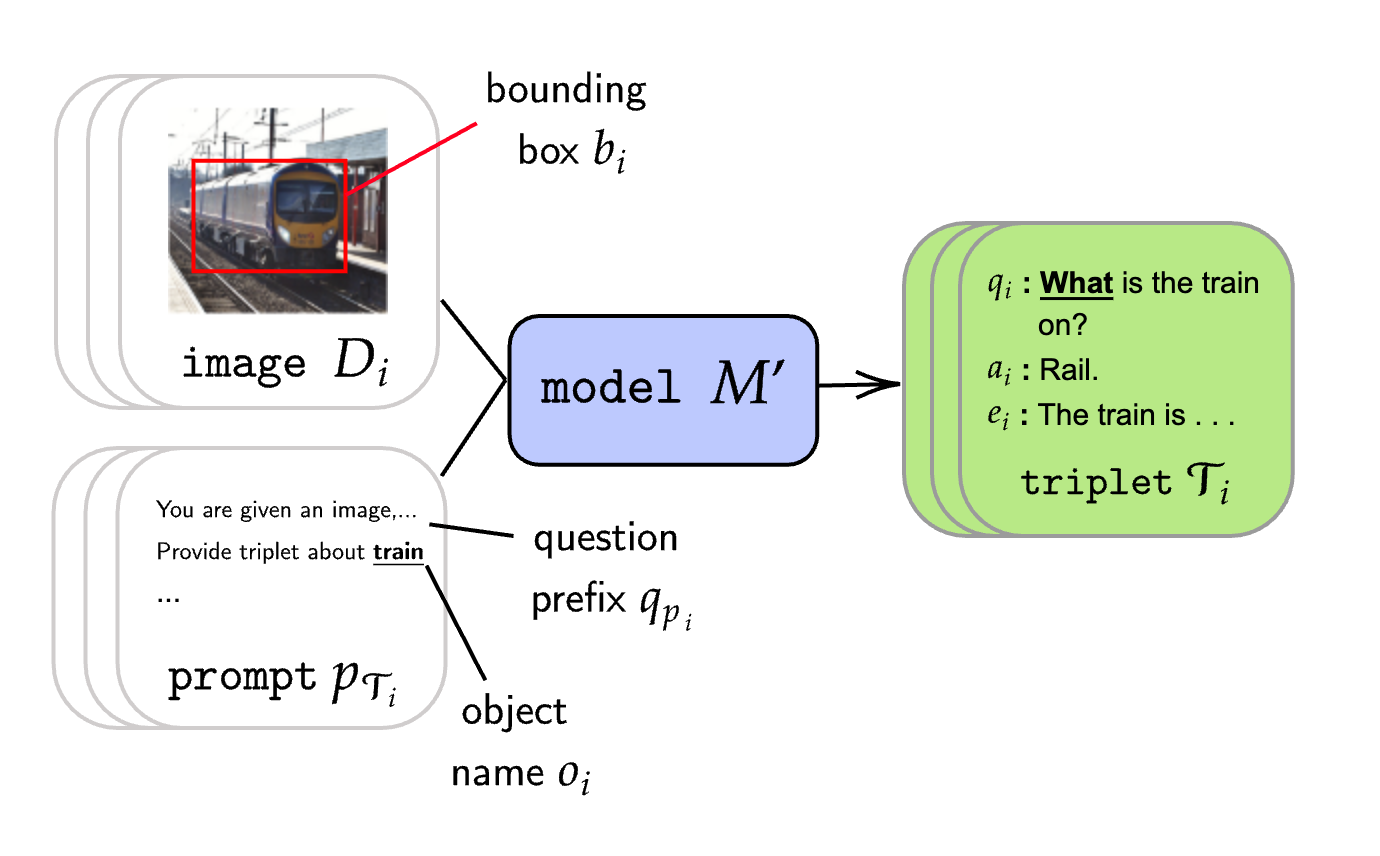} 
    \end{subfigure}
    \begin{subfigure}[t]{.95\textwidth}
        \centering
        \caption{\textsc{Multi-Step}}
        \includegraphics[width=\linewidth]{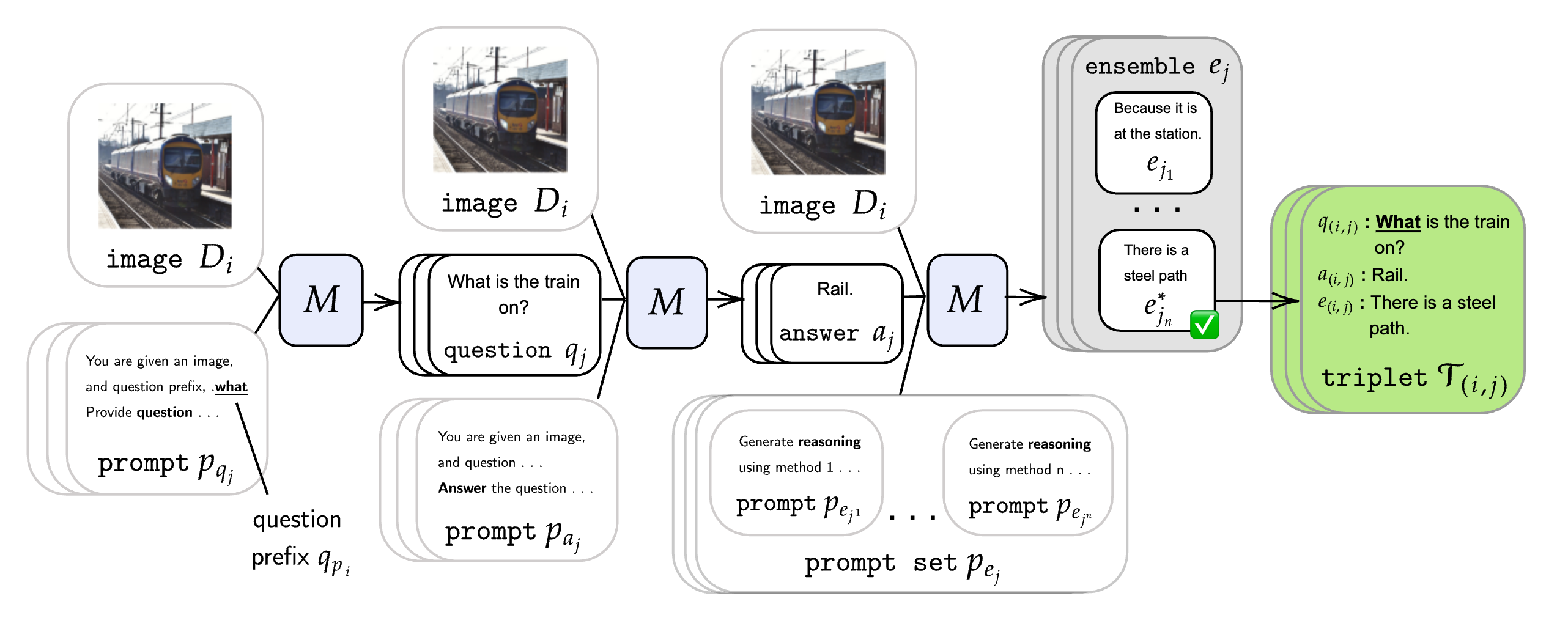} 
    \end{subfigure}
    \caption{An illustrative example of how we construct $\mathcal{T}(q,a,e)$ with three different approaches ($\singlestep$, $\singlestepvip$, and $\multistep$) using model $M$, prompt $p$, and image $D$. Each $D_i$ is used to generate up to $j$ triplets using different formatted prompt $p_{j}$ with supporting instruction (e.g., question prefix).}
    \label{fig:pipeline-viz}
\end{figure*}

In this section, we present a comprehensive description of our synthetic data generation methods, including detailed explanations of the three prompting pipelines employed.
Formally, we define $\mathcal{T}=\{\mathcal{T}_1, \cdots, \mathcal{T}_N\}$, where $\mathcal{T}_i = (q_i, a_i, e_i)$ represent our synthetic data triplet, comprising a question $q_i$, an answer $a_i$, and an explanation $e_i$. We define $P_i = \{p_1, p_2, \dots\}$ as the set of prompts used to generate $\mathcal{T}_i$ with model $M$ and input data $\mathcal{D}_i$. The process of generating each triplet $\mathcal{T}_i$ from a sample $\mathcal{D}_i$ can be expressed using inference denoted by the function $f$ that relates these variables:
\begin{equation}
    \mathcal{T}_i = f(M, \mathcal{D}_i, P_i).
\label{eq:original_formula}
\end{equation}
Next, we employ three distinct prompting pipelines to generate $\mathcal{T}_i$, guided by the formulation in Equation \ref{eq:original_formula}. Each pipeline involves a two-step process: First, we produce contextually relevant $q_i$ and $a_i$ from the input data $\mathcal{D}_i$. Second, we generate an explanation $e_i$ to justify the answer $a$ to the corresponding question $q_i$. The distinctiveness of each pipeline is attributed to the specific prompting and pre-processing techniques used, as illustrated in Figure \ref{fig:pipeline-viz} and elaborated in the following explanations.

\subsection{\textsc{Single-Step}}
We generate synthetic data using a straightforward prompting technique. Initially, we craft a single prompt template $p_i$ to produce $\mathcal{T}_i$ in a single inference step. Next, we create a question prefix pool using stratified sampling based on the desired prefixes and their respective proportions, specified as hyper-parameters. We then select a sampled question prefix $q_{p_i}$ and incorporate it into the prompt via a formatting process, denoted as $\Phi(p_i, q_{p_i})$. Hence, we can revise Equation \ref{eq:original_formula} to represent this $\singlestep$ instruction pre-processing as:
\begin{equation}
    \mathcal{T}_i = f(M, \mathcal{D}_i, \Phi(p_i,q_{p_i})).
    \label{eq:naive}
\end{equation}
The details for $\singlestep$ approach and the corresponding prompt template for this method are provided in Appendix \ref{eq:appendix-pseudo-naive}.

\subsection{\singlestepvip }
Several studies have explored the idea of enhancing image interpretation by considering the objects they contain. \citet{zhu2016visual7wgroundedquestionanswering} establish connections between object regions and textual responses, while \citet{lovenia2023negativeobjectpresenceevaluation} introduce a novel approach by utilizing object lists to regenerate question templates. With the emergence of visual-prompt-aware models, such as \citet{cai2024vipllavamakinglargemultimodal}, we aim to employ a similar strategy to enhance the relevance and quality of $\mathcal{T}$ through the integration of regional subcontext extracted from image scene graphs.

At a high level, we follow the $\singlestep$ approach for generating $\mathcal{T}$ with minor adaptation. This adjustments involve the following steps: First, we annotate the dataset $D$ with bounding box $b$ as our additional visual prompt (ViP), denoted as $\Theta(D,b)$. Second, we format $p$ with not only question prefix $q_p$, but also object name $o$ as additional instruction contexts to guide the inference towards the relevant object, denoted as $\Phi(p,q_p,o)$. Lastly, we employ an instruction-tuned model $M'$ trained for visual prompts, instead of the base model $M$. Thus, we showcase this visual prompt fusion by modifying Eq. \ref{eq:naive} as follows:
\begin{equation}
    \mathcal{T} = f(M', \Theta(D,b), \Phi(p,q_p,o)).
    \label{eq:naive_viz}
\end{equation}
The detailed pseudocode for $\singlestepvip$ approach and the corresponding prompts are available in Appendix \ref{eq:appendix-pseudo-naive-sg}.

\subsection{\multistep}
We generate $q_i$, $a_i$, and $e_i$ sequentially rather the previous method's exhaustive generation of all $\mathcal{T}_i$ components in a single step. We aim to enhance the $e$ component by adopting lightweight re-ranking self-consistency \cite{jain2024lightweight}, a framework that enables the ensemble of open-ended generation by selecting the $i$-th generation with the highest average fractional generalized similarity score ${GSC}_{Sim}$ among the other $K-1$ outputs, denoted as ${GSC}_{Sim}(i) = \frac{1}{K-1} \sum_{\substack{j=1,j \neq i}}^{K} Sim(i, j)$. We change the similarity function $Sim$ by utilizing encoder model~\cite{song2020mpnetmaskedpermutedpretraining} instead of the unigram consistency score.

Initially, we design multiple prompt templates, including $p_q$ for question generation, $p_a$ for answer generation, and a set $P_e = \{p_{e_1}, p_{e_2}, ...\}$ consisting of multiple explanation generation instructions, whose outputs are combined using ${GSC}_{Sim}$. We then format each $p$ with its required contexts to construct the subsequent output (e.g., $p_q$ with $q_p$, $p_a$ with $q$, and so forth). Finally, we combine multiple explanations generated from $P_e$, denoted as $\Psi(e_1, e_2, ...)$, to determine the optimal $e^*$ to accompany the previously generated $q$ and $a$. The sequential generation process can be expressed as:
\begin{align}
    q &= f(M, D, \Phi(p_q, q_p)). \\
    a &= f(M, D, \Phi(p_a, q)). \\
    e^* &= \Psi\left( 
        \begin{aligned}
            &f(M, D, \Phi(p_{e_1}, q, a)), \\
            &f(M, D, \Phi(p_{e_2}, q, a)), \\
            &\hspace{1.5cm}\dots
        \end{aligned} 
    \right). \\
    \mathcal{T} &= \left(q, a, e^*\right).
    \label{eq:seq_ensemble}
\end{align}
The details for $\multistep$ and the corresponding prompts are available in Appendix \ref{eq:appendix-pseudo-ensemble}.

\section{Experimental Setup}
In this section, we present the dataset, LVLM, and prompts used to artificially generate our data. These components are used in three experimental settings that are evaluated using the setup to analyze the strengths and limitations of each approach.

\subsection{Dataset} 

We utilize the GQA dataset \cite{GQA} by sampling 10k images from the total collection of 110k along with their associated scene graphs to produce the evaluated data. We ensure that each of the selected images has a corresponding scene graph to facilitate the $\singlestepvip$ setting. Additionally, we apply a filtering criterion to the scene graphs, considering only objects with bounding box areas above a certain area threshold. This filtering step aims to exclude very small and unclear objects, addressing two issues: (1) not all images have corresponding graph objects, and (2) some object graphs have excessively small areas, which could hinder their usefulness during inference.

\subsection{Models} 

We employ three LLaVA-1.5 \cite{liu2024improvedbaselinesvisualinstruction} variants: (a) LLaVA-1.5-7B, (b) LLaVA-1.5-13B, and (c) ViP-LLaVA-13B~\cite{cai2024vipllavamakinglargemultimodal}. The models integrate the pre-trained CLIP ViT-L/14 visual encoder with Vicuna via a simple projection matrix. LVLM (c) is a fine-tuned (b) with annotation-rich image input. The LLaVA with Vicuna backbone is selected due to its superior instruction-following capability compared to other variants, as highlighted in \citet{shao2024visualcotadvancingmultimodal}. 

In the $\singlestep$ setting, we assess performance differences between (a) and (b) to probe the base model ability in following basic instruction to produce our data. We then proceed with (b) to run $\multistep$ setting for more advanced prompting technique. Finally, we employ (c) in conjunction with the boundary box as an additional input in $\singlestepvip$ setting.

\subsection{Prompts} 
We structure our prompt templates by referring to the LLM-as-a-Judge approach, which has demonstrated the effectiveness reference-guided techniques in evaluation tasks \cite{zheng2023judgingllmasajudgemtbenchchatbot}. Our contribution lies in adapting this template for generation purposes, with a specific focus on improving LVLM's rule adherence. We employ a straightforward prompting strategy in the $\singlestep$ setting and make necessary adjustments for additional object names and annotation contexts in the $\singlestepvip$ setting. Furthermore, we explore more advanced prompting methods in the $\multistep$ setting, ensembling multiple reasoning paths such as CoT \cite{wei2023chainofthought} and ReAct \cite{yao2023react}. The prompt templates are provided in Appendix \ref{sec:appendix-prompts}.

\begin{table*}[ht]
\centering
\resizebox{\linewidth}{!}{
\begin{tabular}{lcccc|ccc|ccc}
    \toprule
    \textbf{Model} & \multicolumn{4}{c}{\textbf{Triplet}} & \multicolumn{3}{c}{\textbf{Vocabulary Size}} & \multicolumn{3}{c}{\textbf{Avg. Sentence Length}} \\
    \cmidrule(lr){2-5} \cmidrule(lr){6-8} \cmidrule(lr){9-11}
     & \# Valid Instance & \# Unique & Valid \% & Unique \% & $q$ & $a$ & $e$ & $q$ & $a$ & $e$ \\
    \midrule
    $\singlestep$\textsc{-7B}    & 19,309 & 15,328 & 94.2\% & 79.4\% & 5,196 & 11,495 & 12,645 & 8.7 & 8.0 & 28.1 \\
    $\singlestep$\textsc{-13B}   & 20,501 & 16,847 & 100.0\% & 82.2\% & 6,170 & 10,198 & 11,448 & 8.8 & 4.4 & 22.3 \\
    $\singlestepvip$  & 18,458 & 16,968 & 90.0\% & 91.9\% & 5,437 & 10,066 & 11,717 & 8.5 & 6.2 & 18.1 \\
    $\multistep$  & 20,501 & 17,539 & 100.0\% & 85.6\% & 8,163 & 8,345 & 14,340 & 10.5 & 2.3 & 42.2 \\
    \bottomrule
\end{tabular}
}
\caption{Experiment results across four different settings in a total of 66,682 valid unique instances over 82,004 expected amount. To determine the unique triplets, post-processing is conducted to filter out duplicate triplets, which has the same all three $q$, $a$, and $e$ components. The vocabulary size and sentence length is derived by using a simple tokenization to separate whitespace on the cleaned sampled corpus without punctuation.}
\label{tab:general_stats}
\end{table*}

\subsection{Experiment Settings}
We define our methodology formulation into the following experiment settings as following:
\begin{itemize}
    \setlength\itemsep{0em}
    \item $\singlestep$\textsc{-7B}: We implement the $\singlestep$ method, utilizing the LLaVA-1.5-7B model to assess the smallest model's capability in generating the dataset.
    \item $\singlestep$\textsc{-13B}: This setting is similar to the previous one, but employs the LLaVA-1.5-13B model to investigate the impact of model scale differences on synthetic data quality and similarity.
    \item $\singlestepvip$: We introduce visual-prompt-infused input data with the ViP-LLaVA-13B model as our instruction-tuned model.
    \item \multistep: We apply the sequential ensemble method using the LLaVA-1.5-13B model.
    \item \textsc{Human}: We conduct human annotations with 10 annotators to generate 10 triplets each, enabling a comparison with our synthetic data pipeline.
\end{itemize}
Our goal is to produce 20,501 synthetic triplets for each experiment. The specific hyper-parameters and engine details can be found in Appendix \ref{sec:appendix-hyperparams}.

\subsection{Evaluation Settings}
\label{subsec:triplet-eval}

We evaluate 2,004 out of 82,004 synthetic triplets, comprised of 501 per setting. We then apply post-processing to assess both validity rate and time efficiency ($\Bar{t}$) across all triplets. Finally, we use only the valid triplets from the 501 data points for similarity evaluation and sample 50 of them for quality evaluation.

\paragraph{\textbf{Quality Evaluation}} 
\label{sec:quality-metrics}

Quantifying the quality of synthesized explanations for VQA is challenging, so we opt for a human evaluation approach as our primary evaluation source. Three annotators assess our synthetic data following the guidelines in Appendix \ref{sec:appendix-eval-rules}, and we also calculate the inter-annotator agreement Gwet-AC2 metric to ensure consistency. We employ a human evaluation scoring framework inspired by \citet{zhang2023llmevalpreliminarystudyevaluate}, making slight adjustments to align with our evaluation objectives. We emphasize assessment on \textit{Accuracy}, \textit{Logical}, \textit{Clarity}, \textit{Detail}, and \textit{Relevancy} criteria, with each scored on a scale from 1 (worst) to 3 (best). Details on human annotation metrics are shown in Appendix~\ref{annotation-metrics}. Furthermore, we employ BERTScore, CLIPScore, and ROUGE metrics to support our subjective evaluation in assessing the quality of generated $(q,a)$ and $e$ against one another and the input image.

\begin{figure*}[!th]
    \centering
    \includegraphics[width=0.98\linewidth]{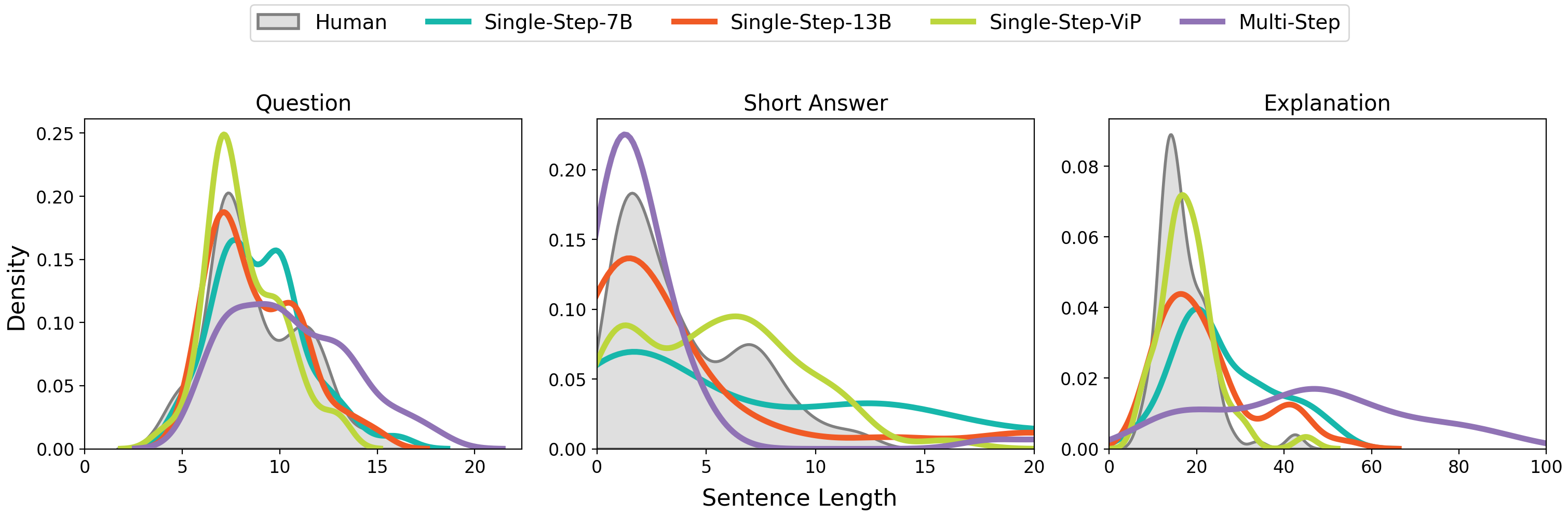} 
    \caption{Comparison of density estimation for sentence length distribution across all experiment settings and human-generated $\mathcal{T}$. It provides a visual representation of the distribution differences, with more detailed numerical insights available in Table \ref{tab:similarity_table}.}
    \label{fig:structure-simil}
    \vspace{-0.5mm}
\end{figure*}

\paragraph{\textbf{Similarity Evaluation}} We perform a comparative analysis by examining the text length distribution variations in $q$, $a$, and $e$ between our synthetic data and human-generated data. To quantify these differences, we utilize Jensen-Shannon Divergence (JSD) and Pearson correlation as our evaluation metrics. JSD is employed to measure distribution shifts within a specific range, while Pearson correlation assesses the overall trend in text length across the two datasets.

\paragraph{\textbf{Validity and Efficiency Evaluation}}We perform a supplementary analysis to evaluate the reliability of each experiment set in generating valid data. Valid data, denoted as $\mathcal{T}_\text{valid}$, is defined as a triplet $\mathcal{T}$ whose elements collectively follow a predefined regular expression, ensuring natural sentence structure. For $\singlestepvip$, valid triplets must also not include any hidden annotation contexts (e.g., \texttt{``...in the red bounding box."}). The validity proportion is defined as $\text{Valid }\% = \frac{{|\mathcal{T}_\text{valid}|}}{|\mathcal{T}_\text{valid}| + {|\mathcal{T}_\text{invalid}|}}  \times 100\%$.
Additionally, we conduct a comparative analysis of the time efficiency of our approach against conventional methodologies. We measure the average time $\bar{t}$ required to generate each valid data point by recording the total time for dataset creation and dividing it by the number of valid triplets, $|\mathcal{T}_\text{valid}|$, denoted as $\bar{t}=\frac{t}{|\mathcal{T}_\text{valid}|}$.

\section{Results and Analysis}
In this section, we present the general data generation and evaluation results for the entire dataset, as shown in Table \ref{tab:general_stats}. We then highlight key findings through quality and similarity analyses, using 50 sampled triplets from each setting. All discussions refer to Table \ref{tab:similarity_table} for similarity evaluation and Table \ref{tab:quality_eval_metrics} for quality evaluation.

\subsection{Experiment Results}
\paragraph{$\singlestep$\textsc{-7B}}
Our first setting successfully generated 94.2\% valid triplets over the expected amount. Furthermore, a decent similarity is observed, with a Pearson correlation of 0.70 and a Jensen-Shannon Divergence (JSD) of 0.35. This similarity is predominantly attributed to the high similarity in the $q$ component, as illustrated in Figure \ref{fig:structure-simil}. However, only 79.4\% of these triplets are unique, indicating a moderate scalability level when this approach is used to generate unique synthetic data. In terms of quality, this approach achieves a score of 2.519, which is 12\% short compared to human-generated data.

\paragraph{$\singlestep$\textsc{-13B}} 
In contrast to the previous setting, the $\singlestep$ setting with a larger model excels by generating 100\% of the expected quantity. These triplets have an uniqueness rate of 82.5\%. The similarity metrics also improve substantially, with a Pearson correlation of 0.80 and a JSD of 0.30. The $q$, $a$, and $e$ components contribute evenly to these scores. In terms of quality, this approach scores 2.619, showing notable improvements over the smaller model in all criteria except for the detail aspect, reducing the gap to human quality to just 8\%.

\paragraph{\multistep} 
The multi-step setting generates 100\% valid triplets with an improved unique rate of 85.9\%. While the overall quality score slightly improves to 2.623, there is a notable drop in relevancy. This tendency for overly detailed explanations is also evident in the similarity metrics, resulting in the lowest performance among all settings, with a Pearson correlation of 0.58 and a JSD of 0.39.

\paragraph{$\singlestepvip$}
The visual-prompt infused $\singlestep$ setting generates 90\% valid triplets, 10\% lower compared to $\multistep$. Despite this, a notable 92.1\% unique rate is achieved indicating that visual-prompt helps to generate unique data instance. This setting also exhibits improvement in similarity, marking Pearson correlation of 0.84 and JSD of only 0.25. Furthermore, it increases quality score to 2.646, lacking only 7\% compared to human. This is predominantly achieved by alleviating the lack of relevancy suffered by previous settings while maintaining decent score of other criteria and championing in clarity aspect over human.

\begin{table}[ht]
\centering
\resizebox{1\columnwidth}{!}{
\begin{tabular}{lccc|c|ccc|c}
\toprule
\textbf{Model} & \multicolumn{4}{c}{\textbf{Pearson}} & \multicolumn{4}{c}{\textbf{JSD}} \\
\cmidrule(lr){2-5} \cmidrule(lr){6-9}
 & \textbf{$q$} & \textbf{$a$} & \textbf{$e$} & \textbf{Avg} & \textbf{$q$} & \textbf{$a$} & \textbf{$e$} & \textbf{Avg} \\
\midrule
$\singlestep$\textsc{-7B}     & 0.88 & 0.70 & 0.50 & 0.70 & 0.16 & 0.43 & 0.44 & 0.35 \\
$\singlestep$\textsc{-13B}    & 0.93 & 0.75 & 0.75 & 0.81 & 0.18 & 0.41 & 0.32 & 0.30 \\
$\singlestepvip$     & 0.96 & 0.74 & 0.83 & \textcolor{ForestGreen}{0.84} & 0.17 & 0.30 & 0.28 & \textcolor{ForestGreen}{0.25} \\
$\multistep$   & 0.78 & 0.76 & 0.22 & \textcolor{red!70}{0.58} & 0.27 & 0.43 & 0.47 & \textcolor{red!70}{0.39} \\
\bottomrule
\end{tabular}%
}
\caption{Similarity metrics evaluation result for different experiments, including averages for Pearson correlation and Jensen-Shannon Divergence. The evaluation is conducted to the 501 sampled synthetic data from each experiment setting compared to human generated data.}
\label{tab:similarity_table}
\end{table}

\begin{table*}[ht]
\centering
\resizebox{\linewidth}{!}{
\begin{tabular}{lcccccc|cc|c|c}
    \toprule
    \textbf{Model} & \multicolumn{6}{c}{\textbf{Human Evaluation}} & \multicolumn{2}{c}{\textbf{CLIPScore}} & \textbf{BERTScore} & \textbf{ROUGE} \\
    \cmidrule(lr){2-7} \cmidrule(lr){8-9} \cmidrule(lr){10-10} \cmidrule(lr){11-11}
     & Accuracy & Logic & Clarity & Detail & Relevancy& \textbf{Avg. Score} & $q,a \leftrightarrow D$ & $e \leftrightarrow D$ & $q,a \leftrightarrow e$ & $q,a \leftrightarrow e$  \\
    \midrule
    \textsc{Human}            & 2.82 & 2.79 & 2.86 & 2.93 & 2.88 & 2.86 & 23.33 & 23.13 & 0.745 & 0.45 \\
    \midrule
    $\singlestep$\textsc{-7B}    & \textcolor{red!70}{2.56} & \textcolor{red!70}{2.47} & \textcolor{red!70}{2.74} & 2.52 & \textcolor{red!70}{2.30} & \textbf{\textcolor{red!70}{2.519}} & 28.49 & 30.20 & 0.772 & 0.34 \\
    $\singlestep$\textsc{-13B}   & \textcolor{ForestGreen}{2.71} & 2.51 & 2.86 & \textcolor{red!70}{2.44} & \underline{2.57} & \textbf{2.619} & 28.38 & \textcolor{ForestGreen}{30.66} & \textcolor{ForestGreen}{0.779} & \textcolor{ForestGreen}{0.40} \\
    $\singlestepvip$      & 2.57 & \underline{2.53} & \textcolor{ForestGreen}{2.91} & \underline{2.59} & \textcolor{ForestGreen}{2.63} & \textbf{\textcolor{ForestGreen}{2.646}} & \textcolor{red!70}{24.42} & \textcolor{red!70}{25.62} & 0.758 & 0.37 \\
    $\multistep$   & \underline{2.69} & \textcolor{ForestGreen}{2.64} & 2.76 & \textcolor{ForestGreen}{2.68} & 2.35 & \underline{\textbf{2.623}} & \textcolor{ForestGreen}{28.71} & 27.77 & \textcolor{red!70}{0.749} & \textcolor{red!70}{0.30} \\
    \bottomrule
\end{tabular}
}
\caption{Quality evaluation results are obtained through human and automated metrics, utilizing 100 surveyed data points from human sources and 50 sampled synthetic data points across all experiment settings. We follow the human evaluation criteria outlined in Section \ref{subsec:triplet-eval}, where \textcolor{ForestGreen}{green}, \underline{underlined}, and \textcolor{red!70}{red} numbers denote the best, second-best, and worst results, respectively. In automated evaluation, $x \leftrightarrow y$ denotes the score calculation based on the interaction between $x$ and $y$.}
\label{tab:quality_eval_metrics}
\end{table*}

\subsection{Larger Model Improves Instruction Obedience and Generated Data Quality}
\label{subsec:larger-improve-quality}

Table \ref{tab:quality_eval_metrics} reveals that $\singlestep$\textsc{-7B} achieves the lowest scores in 4 out of 5 human evaluation criteria, resulting in the worst overall performance among all settings. By using a larger model variant, $\singlestep$\textsc{-13B} secures a 5\% overall improvement, including a significant 12\% boost in relevancy, without any instruction modification. Other improvements are also evident in the distribution similarity. Figure \ref{fig:structure-simil} illustrates a better data distribution spread in $\singlestep$\textsc{-13B} compared to $\singlestep$\textsc{-7B}, closely resembling the human-generated data distribution. This improvement is also evidenced by a 14\% increase in Pearson correlation and a 13\% reduction in JSD distance, as shown in Table \ref{tab:similarity_table}. These findings highlight that the size of LLaVA-1.5 scales positively with the quality \& similarity metrics improvement, leveraging the advantages of larger models: (1) better instruction following ability, resulting in reduced formatting errors and better text length control, and (2) generation of more logically relevant explanations, leading to high-quality data.

\subsection{Irrelevant Logical Results in \multistep}
\label{subsec:irrelevant-logical-res}
The ensembling method effectively enhances the logic criterion, outperforming all other settings. This suggests that employing advanced prompting techniques such as CoT and ReAct can significantly boost LLaVA's reasoning abilities, resulting in more detailed and logically sound explanations. However, this improvement is accompanied by a trade-off, as the relevancy criterion score decreases by 9\%. This finding indicates that while longer and more elaborate explanations may enhance logical coherence, they can compromise explanation precision. This observation is further substantiated by Figure \ref{fig:structure-simil}, which illustrates a notable shift in the explanation distribution compared to other settings. 

We attribute this issue to the chosen explanation sourced from either CoT or ReAct output prompts. As detailed in Appendix \ref{sec:appendix-prompts}, both techniques involve generating more detailed intermediary steps to reach the final conclusion. This can potentially lead to overly long outputs from the LLM, resulting in detailed but less precise explanations. These findings emphasize the importance of (1) implementing an appropriate token-limiting mechanism and (2) exploring sequential intermediary step generation to prevent overly large outputs and maintain explanation precision.

\subsection{Overcoming Relevancy Issues with Visual Prompts}
\label{subsec:overcoming-relevancy-visprompt}

Enhancing the $\singlestep$ method with visual prompts, particularly bounding boxes, results in a significant 12\% increase in the relevancy score compared to the $\multistep$ setting. Interestingly, all other criteria maintain impressive scores, including clarity, which surpasses human explanation quality. Furthermore, similarity evaluation reveals that $\singlestepvip$ achieves the best similarity score, with a Pearson correlation of 0.84 and a minimized JSD of 0.25. This exceptional performance is notably attributed to how the explanation aspect closely follows the human-generated data distribution as depicted in Figure \ref{fig:structure-simil}. These findings highlight that visual prompts serve as an effective guide for ViP-LLaVA, enabling it to produce logically relevant explanations while maintaining excellence in other aspects.

\subsection{Automatic Quality Metrics Analysis}

Recall that we employ BERTScore, CLIPScore, and ROUGE as our automated quality metrics in Table \ref{tab:quality_eval_metrics}. While we do not explore into every detail, we highlight some interesting points that reinforce our previous analyses. The enhanced performance across all automated metrics indicates that $\singlestep$\textsc{-13B} surpasses $\singlestep$\textsc{-7B}, consistently delivering superior results in all $\mathcal{T}$ components and reinforcing the analysis from Section \ref{subsec:larger-improve-quality}. Secondly, the lowest scores observed for BERTScore and ROUGE in the $\multistep$ setting validate the presence of irrelevant context within $e$, as elaborated in Section \ref{subsec:irrelevant-logical-res}. This observation highlights the diminished contextual relevance between $q, a$ and $e$. Lastly, the worst result for CLIPScore in the $\singlestepvip$ setting, which contrasts the human evaluation's best mark, can be attributed to the localization technique detailed in Section \ref{subsec:overcoming-relevancy-visprompt}. It is important to note that CLIPScore is an image captioning metric, designed to assess how well a text contextually represents the entire image. In contrast, $\singlestepvip$ captures subregions of the image to increase focus on specific contexts by utilizing bounding box. Given that both $q,a$, and $e$ are more likely to represent subregions rather than the entire image, this discrepancy leads to the observed lower CLIPScore performance.

\subsection{Time Efficiency}

Table \ref{tab:efficiency-rate} presents the inference times required to generate 501 synthetic data points for each experiment. The results demonstrate that generating triplets $\mathcal{T}$ synthetically can achieve up to a 20$\times$ efficiency improvement compared to the human generation approach. All $\singlestep$-based experiments yield comparable results, while $\multistep$ is approximately four times slower due to its sequential generation and multiple reasoning paths.
\begin{table}[!ht]
\centering
\resizebox{0.49\textwidth}{!}{
    \begin{tabular}{lrrr}
    \toprule
    \textbf{Model} & \textbf{$|{\mathcal{T}_\text{valid}}|$} & \textbf{$t$} & \textbf{$\Bar{t}$} \\
    \midrule
    Human & 501$^*$ & 350m 5s$^*$ & 42.1s (1$\times$)\\
     \midrule
    $\singlestep$\textsc{-7B} & 476 & \underline{16m 41s} & \textbf{2.10s} (20.0$\times$) \\
    $\singlestep$\textsc{-13B} & 501 & 19m 23s & 2.32s (18.1$\times$) \\
    $\singlestepvip$ & 450 & \textbf{15m 54s} & \underline{2.12s} (19.9$\times$) \\
    $\multistep$ & 483 & 66m 51s & 8.01s (5.3$\times$) \\
    \bottomrule
    \end{tabular}
}
\caption{Comparison of synthetic data generation time for all experiments. The multiplier for $\Bar{t}$ is computed relative to the human annotation time ($^*$estimated time per 501 generated data), following formula in section \ref{subsec:triplet-eval}. \textbf{Bold} and \underline{underlined} times represent the most efficient and second-most efficient experiments, respectively.}
\label{tab:efficiency-rate}
\end{table}

\section{Related Work}

\paragraph{Explanation Generation in VQA} Several studies have emphasized the generation of explanations, either manually or automatically. Manual approaches, such as VCR \cite{zellers2019recognitioncognitionvisualcommonsense} and e-SNLI-VE \cite{do2021esnlivecorrectedvisualtextualentailment}, employ human annotators to derive explanations from existing VQA datasets. In contrast, automatic methods like GQA-REX \cite{chen2022rexreasoningawaregroundedexplanation} utilize functional programming, allowing automatic explanations generation which are grounded on the reasoning process and tightly couple keywords and regions of interest. Another automatic methods like VQA-E \cite{li2018vqaeexplainingelaboratingenhancing} aligns and merges constituency parse trees from QA-caption pairs, while VQA-X \cite{park2018multimodalexplanationsjustifyingdecisions} employs separate answer and explanation models for generation. In this paper, we introduce a novel approach by proposing a unified model using single LVLM, eliminating multiple architectures need.

\paragraph{Neural Synthetic Data Generation} In the realm of multi-modal learning, particularly in the vision-language domain, the potential of synthetic data generation has been extensively explored. \citet{li2023syntheticdatagenerationlarge} discuss the application of synthetic data across various tasks and modalities. In computer vision, GAN-based models \cite{karras2019stylebasedgeneratorarchitecturegenerative} and diffusion-based approaches \cite{nichol2022glidephotorealisticimagegeneration} are utilized for image synthesis. Within natural language processing domain, several studies \cite{kumar2021dataaugmentationusingpretrained, chung_2023, schmidt2024promptingbasedsyntheticdatageneration} employ synthetic data to enhance in-context learning. In the joint vision-language training, \citet{multimodalcaptioningxiao2023} leverage diffusion models to generate image captions, and \citet{lovenia2023negativeobjectpresenceevaluation} create intermediary data to support object hallucination analysis. The joint training aligns vision and language representations, resulting in more relevant generation~\citep{winata2024preference}. In this work, we primarily focus on utilizing LVLMs to evaluate their capability in generating high-quality VQA-NLE data.

\section{Conclusion}
In this paper, we propose efficient methods for VQA-NLE data generation that leverage LVLMs through single-step and multi-step pipelines. Our methods produce high-quality data, achieving up to 20$\times$ greater time efficiency compared to traditional human annotation, with only a slight reduction in quality that remains closely comparable to human-annotated data without further fine-tuning. We demonstrate that incorporating visual prompts significantly enhances the relevance of text generation. Additionally, we emphasize the scalability of our approach to showcase the robustness of our solution for automatically generating multi-modal NLE data on an even larger scale.

\section*{Limitations}
We have identified several promising avenues for enhancing our research outcomes. However, for the scope of this study, we have chosen to limit our experiments to three distinct vision-language pre-trained models. This focused approach allows us to conduct a more detailed and manageable analysis within the constraints of our current resources and time frame. By concentrating on these specific models, we aim to provide a thorough evaluation and establish a solid foundation for future research.

\section*{Ethical Considerations}
Our research focuses on generating synthetic VQA-NLE. We are committed to conducting our evaluations with the highest standards of transparency and fairness. Additionally, we ensure that the generation process strictly excludes any sensitive or personal data.



\bibliography{ref}

\begin{thebibliography}{38}
\providecommand{\natexlab}[1]{#1}

\bibitem[{Aggarwal et~al.(2021)Aggarwal, Mandowara, Agrawal, Khandelwal, Singla, and Garg}]{aggarwal-etal-2021-explanations}
Shourya Aggarwal, Divyanshu Mandowara, Vishwajeet Agrawal, Dinesh Khandelwal, Parag Singla, and Dinesh Garg. 2021.
\newblock \href {https://doi.org/10.18653/v1/2021.acl-long.238} {{E}xplanations for {C}ommonsense{QA}: {N}ew {D}ataset and {M}odels}.
\newblock In \emph{Proceedings of the 59th Annual Meeting of the Association for Computational Linguistics and the 11th International Joint Conference on Natural Language Processing (Volume 1: Long Papers)}, pages 3050--3065, Online. Association for Computational Linguistics.

\bibitem[{Bai et~al.(2023)Bai, Bai, Yang, Wang, Tan, Wang, Lin, Zhou, and Zhou}]{bai2023qwenvlversatilevisionlanguagemodel}
Jinze Bai, Shuai Bai, Shusheng Yang, Shijie Wang, Sinan Tan, Peng Wang, Junyang Lin, Chang Zhou, and Jingren Zhou. 2023.
\newblock \href {https://arxiv.org/abs/2308.12966} {Qwen-vl: A versatile vision-language model for understanding, localization, text reading, and beyond}.
\newblock \emph{Preprint}, arXiv:2308.12966.

\bibitem[{Cai et~al.(2024)Cai, Liu, Park, Mustikovela, Meyer, Chai, and Lee}]{cai2024vipllavamakinglargemultimodal}
Mu~Cai, Haotian Liu, Dennis Park, Siva~Karthik Mustikovela, Gregory~P. Meyer, Yuning Chai, and Yong~Jae Lee. 2024.
\newblock \href {https://arxiv.org/abs/2312.00784} {Vip-llava: Making large multimodal models understand arbitrary visual prompts}.
\newblock \emph{Preprint}, arXiv:2312.00784.

\bibitem[{Chen and Zhao(2022)}]{chen2022rexreasoningawaregroundedexplanation}
Shi Chen and Qi~Zhao. 2022.
\newblock \href {https://arxiv.org/abs/2203.06107} {Rex: Reasoning-aware and grounded explanation}.
\newblock \emph{Preprint}, arXiv:2203.06107.

\bibitem[{Chung et~al.(2023)Chung, Kamar, and Amershi}]{chung_2023}
John Chung, Ece Kamar, and Saleema Amershi. 2023.
\newblock \href {https://doi.org/10.18653/v1/2023.acl-long.34} {Increasing diversity while maintaining accuracy: Text data generation with large language models and human interventions}.
\newblock In \emph{Proceedings of the 61st Annual Meeting of the Association for Computational Linguistics (Volume 1: Long Papers)}. Association for Computational Linguistics.

\bibitem[{Do et~al.(2021)Do, Camburu, Akata, and Lukasiewicz}]{do2021esnlivecorrectedvisualtextualentailment}
Virginie Do, Oana-Maria Camburu, Zeynep Akata, and Thomas Lukasiewicz. 2021.
\newblock \href {https://arxiv.org/abs/2004.03744} {e-snli-ve: Corrected visual-textual entailment with natural language explanations}.
\newblock \emph{Preprint}, arXiv:2004.03744.

\bibitem[{Hendricks et~al.(2016)Hendricks, Akata, Rohrbach, Donahue, Schiele, and Darrell}]{hendricks2016generatingvisualexplanations}
Lisa~Anne Hendricks, Zeynep Akata, Marcus Rohrbach, Jeff Donahue, Bernt Schiele, and Trevor Darrell. 2016.
\newblock \href {https://arxiv.org/abs/1603.08507} {Generating visual explanations}.
\newblock \emph{Preprint}, arXiv:1603.08507.

\bibitem[{Hudson and Manning(2019)}]{GQA}
Drew~A. Hudson and Christopher~D. Manning. 2019.
\newblock \href {https://arxiv.org/abs/1902.09506} {{GQA:} a new dataset for compositional question answering over real-world images}.
\newblock \emph{CoRR}, abs/1902.09506.

\bibitem[{Jain et~al.(2024)Jain, Ma, Deoras, and Xiang}]{jain2024lightweight}
Siddhartha Jain, Xiaofei Ma, Anoop Deoras, and Bing Xiang. 2024.
\newblock \href {https://arxiv.org/abs/2307.06857} {Lightweight reranking for language model generations}.
\newblock \emph{Preprint}, arXiv:2307.06857.

\bibitem[{Karras et~al.(2019)Karras, Laine, and Aila}]{karras2019stylebasedgeneratorarchitecturegenerative}
Tero Karras, Samuli Laine, and Timo Aila. 2019.
\newblock \href {https://arxiv.org/abs/1812.04948} {A style-based generator architecture for generative adversarial networks}.
\newblock \emph{Preprint}, arXiv:1812.04948.

\bibitem[{Kayser et~al.(2021)Kayser, Camburu, Salewski, Emde, Do, Akata, and Lukasiewicz}]{kayser2021vil}
Maxime Kayser, Oana-Maria Camburu, Leonard Salewski, Cornelius Emde, Virginie Do, Zeynep Akata, and Thomas Lukasiewicz. 2021.
\newblock e-vil: A dataset and benchmark for natural language explanations in vision-language tasks.
\newblock In \emph{Proceedings of the IEEE/CVF international conference on computer vision}, pages 1244--1254.

\bibitem[{Kayser et~al.(2022)Kayser, Emde, Camburu, Parsons, Papiez, and Lukasiewicz}]{kayser2022explainingchestxraypathologies}
Maxime Kayser, Cornelius Emde, Oana-Maria Camburu, Guy Parsons, Bartlomiej Papiez, and Thomas Lukasiewicz. 2022.
\newblock \href {https://arxiv.org/abs/2207.04343} {Explaining chest x-ray pathologies in natural language}.
\newblock \emph{Preprint}, arXiv:2207.04343.

\bibitem[{Kim et~al.(2018)Kim, Rohrbach, Darrell, Canny, and Akata}]{kim2018textualexplanationsselfdrivingvehicles}
Jinkyu Kim, Anna Rohrbach, Trevor Darrell, John Canny, and Zeynep Akata. 2018.
\newblock \href {https://arxiv.org/abs/1807.11546} {Textual explanations for self-driving vehicles}.
\newblock \emph{Preprint}, arXiv:1807.11546.

\bibitem[{Kotonya and Toni(2020)}]{kotonya2020explainableautomatedfactcheckingsurvey}
Neema Kotonya and Francesca Toni. 2020.
\newblock \href {https://arxiv.org/abs/2011.03870} {Explainable automated fact-checking: A survey}.
\newblock \emph{Preprint}, arXiv:2011.03870.

\bibitem[{Kumar et~al.(2021)Kumar, Choudhary, and Cho}]{kumar2021dataaugmentationusingpretrained}
Varun Kumar, Ashutosh Choudhary, and Eunah Cho. 2021.
\newblock \href {https://arxiv.org/abs/2003.02245} {Data augmentation using pre-trained transformer models}.
\newblock \emph{Preprint}, arXiv:2003.02245.

\bibitem[{Li et~al.(2018)Li, Tao, Joty, Cai, and Luo}]{li2018vqaeexplainingelaboratingenhancing}
Qing Li, Qingyi Tao, Shafiq Joty, Jianfei Cai, and Jiebo Luo. 2018.
\newblock \href {https://arxiv.org/abs/1803.07464} {Vqa-e: Explaining, elaborating, and enhancing your answers for visual questions}.
\newblock \emph{Preprint}, arXiv:1803.07464.

\bibitem[{Li et~al.(2023)Li, Zhu, Lu, and Yin}]{li2023syntheticdatagenerationlarge}
Zhuoyan Li, Hangxiao Zhu, Zhuoran Lu, and Ming Yin. 2023.
\newblock \href {https://arxiv.org/abs/2310.07849} {Synthetic data generation with large language models for text classification: Potential and limitations}.
\newblock \emph{Preprint}, arXiv:2310.07849.

\bibitem[{Ling et~al.(2017)Ling, Yogatama, Dyer, and Blunsom}]{ling-etal-2017-program}
Wang Ling, Dani Yogatama, Chris Dyer, and Phil Blunsom. 2017.
\newblock \href {https://doi.org/10.18653/v1/P17-1015} {Program induction by rationale generation: Learning to solve and explain algebraic word problems}.
\newblock In \emph{Proceedings of the 55th Annual Meeting of the Association for Computational Linguistics (Volume 1: Long Papers)}, pages 158--167, Vancouver, Canada. Association for Computational Linguistics.

\bibitem[{Liu et~al.(2024)Liu, Li, Li, and Lee}]{liu2024improvedbaselinesvisualinstruction}
Haotian Liu, Chunyuan Li, Yuheng Li, and Yong~Jae Lee. 2024.
\newblock \href {https://arxiv.org/abs/2310.03744} {Improved baselines with visual instruction tuning}.
\newblock \emph{Preprint}, arXiv:2310.03744.

\bibitem[{Lovenia et~al.(2023)Lovenia, Dai, Cahyawijaya, Ji, and Fung}]{lovenia2023negativeobjectpresenceevaluation}
Holy Lovenia, Wenliang Dai, Samuel Cahyawijaya, Ziwei Ji, and Pascale Fung. 2023.
\newblock \href {https://arxiv.org/abs/2310.05338} {Negative object presence evaluation (nope) to measure object hallucination in vision-language models}.
\newblock \emph{Preprint}, arXiv:2310.05338.

\bibitem[{Lu et~al.(2022)Lu, Mishra, Xia, Qiu, Chang, Zhu, Tafjord, Clark, and Kalyan}]{lu2022learnexplainmultimodalreasoning}
Pan Lu, Swaroop Mishra, Tony Xia, Liang Qiu, Kai-Wei Chang, Song-Chun Zhu, Oyvind Tafjord, Peter Clark, and Ashwin Kalyan. 2022.
\newblock \href {https://arxiv.org/abs/2209.09513} {Learn to explain: Multimodal reasoning via thought chains for science question answering}.
\newblock \emph{Preprint}, arXiv:2209.09513.

\bibitem[{Lu et~al.(2024)Lu, Shen, Wang, Wang, van Rechem, Fu, and Wei}]{lu2024machinelearningsyntheticdata}
Yingzhou Lu, Minjie Shen, Huazheng Wang, Xiao Wang, Capucine van Rechem, Tianfan Fu, and Wenqi Wei. 2024.
\newblock \href {https://arxiv.org/abs/2302.04062} {Machine learning for synthetic data generation: A review}.
\newblock \emph{Preprint}, arXiv:2302.04062.

\bibitem[{Nichol et~al.(2022)Nichol, Dhariwal, Ramesh, Shyam, Mishkin, McGrew, Sutskever, and Chen}]{nichol2022glidephotorealisticimagegeneration}
Alex Nichol, Prafulla Dhariwal, Aditya Ramesh, Pranav Shyam, Pamela Mishkin, Bob McGrew, Ilya Sutskever, and Mark Chen. 2022.
\newblock \href {https://arxiv.org/abs/2112.10741} {Glide: Towards photorealistic image generation and editing with text-guided diffusion models}.
\newblock \emph{Preprint}, arXiv:2112.10741.

\bibitem[{Park et~al.(2018)Park, Hendricks, Akata, Rohrbach, Schiele, Darrell, and Rohrbach}]{park2018multimodalexplanationsjustifyingdecisions}
Dong~Huk Park, Lisa~Anne Hendricks, Zeynep Akata, Anna Rohrbach, Bernt Schiele, Trevor Darrell, and Marcus Rohrbach. 2018.
\newblock \href {https://arxiv.org/abs/1802.08129} {Multimodal explanations: Justifying decisions and pointing to the evidence}.
\newblock \emph{Preprint}, arXiv:1802.08129.

\bibitem[{Schmidt et~al.(2024)Schmidt, Bartezzaghi, and Vu}]{schmidt2024promptingbasedsyntheticdatageneration}
Maximilian Schmidt, Andrea Bartezzaghi, and Ngoc~Thang Vu. 2024.
\newblock \href {https://arxiv.org/abs/2405.09335} {Prompting-based synthetic data generation for few-shot question answering}.
\newblock \emph{Preprint}, arXiv:2405.09335.

\bibitem[{Shao et~al.(2024)Shao, Qian, Xiao, Song, Zong, Wang, Liu, and Li}]{shao2024visualcotadvancingmultimodal}
Hao Shao, Shengju Qian, Han Xiao, Guanglu Song, Zhuofan Zong, Letian Wang, Yu~Liu, and Hongsheng Li. 2024.
\newblock \href {https://arxiv.org/abs/2403.16999} {Visual cot: Advancing multi-modal language models with a comprehensive dataset and benchmark for chain-of-thought reasoning}.
\newblock \emph{Preprint}, arXiv:2403.16999.

\bibitem[{Song et~al.(2020)Song, Tan, Qin, Lu, and Liu}]{song2020mpnetmaskedpermutedpretraining}
Kaitao Song, Xu~Tan, Tao Qin, Jianfeng Lu, and Tie-Yan Liu. 2020.
\newblock \href {https://arxiv.org/abs/2004.09297} {Mpnet: Masked and permuted pre-training for language understanding}.
\newblock \emph{Preprint}, arXiv:2004.09297.

\bibitem[{Wei et~al.(2023)Wei, Wang, Schuurmans, Bosma, Ichter, Xia, Chi, Le, and Zhou}]{wei2023chainofthought}
Jason Wei, Xuezhi Wang, Dale Schuurmans, Maarten Bosma, Brian Ichter, Fei Xia, Ed~Chi, Quoc Le, and Denny Zhou. 2023.
\newblock \href {https://arxiv.org/abs/2201.11903} {Chain-of-thought prompting elicits reasoning in large language models}.
\newblock \emph{Preprint}, arXiv:2201.11903.

\bibitem[{Winata et~al.(2024)Winata, Zhao, Das, Tang, Yao, Zhang, and Sahu}]{winata2024preference}
Genta~Indra Winata, Hanyang Zhao, Anirban Das, Wenpin Tang, David~D Yao, Shi-Xiong Zhang, and Sambit Sahu. 2024.
\newblock Preference tuning with human feedback on language, speech, and vision tasks: A survey.
\newblock \emph{arXiv preprint arXiv:2409.11564}.

\bibitem[{Xiao et~al.(2023)Xiao, Xu, and Zhang}]{multimodalcaptioningxiao2023}
Changrong Xiao, Sean~Xin Xu, and Kunpeng Zhang. 2023.
\newblock \href {https://doi.org/10.1145/3607827.3616839} {Multimodal data augmentation for image captioning using diffusion models}.
\newblock In \emph{Proceedings of the 1st Workshop on Large Generative Models Meet Multimodal Applications}, MM ’23. ACM.

\bibitem[{Yang et~al.(2015)Yang, Yih, and Meek}]{yang-etal-2015-wikiqa}
Yi~Yang, Wen-tau Yih, and Christopher Meek. 2015.
\newblock \href {https://doi.org/10.18653/v1/D15-1237} {{W}iki{QA}: A challenge dataset for open-domain question answering}.
\newblock In \emph{Proceedings of the 2015 Conference on Empirical Methods in Natural Language Processing}, pages 2013--2018, Lisbon, Portugal. Association for Computational Linguistics.

\bibitem[{Yao et~al.(2023)Yao, Zhao, Yu, Du, Shafran, Narasimhan, and Cao}]{yao2023react}
Shunyu Yao, Jeffrey Zhao, Dian Yu, Nan Du, Izhak Shafran, Karthik Narasimhan, and Yuan Cao. 2023.
\newblock \href {https://arxiv.org/abs/2210.03629} {React: Synergizing reasoning and acting in language models}.
\newblock \emph{Preprint}, arXiv:2210.03629.

\bibitem[{Zellers et~al.(2019{\natexlab{a}})Zellers, Bisk, Farhadi, and Choi}]{zellers2019recognition}
Rowan Zellers, Yonatan Bisk, Ali Farhadi, and Yejin Choi. 2019{\natexlab{a}}.
\newblock From recognition to cognition: Visual commonsense reasoning.
\newblock In \emph{Proceedings of the IEEE/CVF conference on computer vision and pattern recognition}, pages 6720--6731.

\bibitem[{Zellers et~al.(2019{\natexlab{b}})Zellers, Bisk, Farhadi, and Choi}]{zellers2019recognitioncognitionvisualcommonsense}
Rowan Zellers, Yonatan Bisk, Ali Farhadi, and Yejin Choi. 2019{\natexlab{b}}.
\newblock \href {https://arxiv.org/abs/1811.10830} {From recognition to cognition: Visual commonsense reasoning}.
\newblock \emph{Preprint}, arXiv:1811.10830.

\bibitem[{Zhang et~al.(2023)Zhang, Zhang, Yuan, Liu, Shi, Gui, Zhang, and Huang}]{zhang2023llmevalpreliminarystudyevaluate}
Yue Zhang, Ming Zhang, Haipeng Yuan, Shichun Liu, Yongyao Shi, Tao Gui, Qi~Zhang, and Xuanjing Huang. 2023.
\newblock \href {https://arxiv.org/abs/2312.07398} {Llmeval: A preliminary study on how to evaluate large language models}.
\newblock \emph{Preprint}, arXiv:2312.07398.

\bibitem[{Zheng et~al.(2023)Zheng, Chiang, Sheng, Zhuang, Wu, Zhuang, Lin, Li, Li, Xing, Zhang, Gonzalez, and Stoica}]{zheng2023judgingllmasajudgemtbenchchatbot}
Lianmin Zheng, Wei-Lin Chiang, Ying Sheng, Siyuan Zhuang, Zhanghao Wu, Yonghao Zhuang, Zi~Lin, Zhuohan Li, Dacheng Li, Eric~P. Xing, Hao Zhang, Joseph~E. Gonzalez, and Ion Stoica. 2023.
\newblock \href {https://arxiv.org/abs/2306.05685} {Judging llm-as-a-judge with mt-bench and chatbot arena}.
\newblock \emph{Preprint}, arXiv:2306.05685.

\bibitem[{Zhu et~al.(2023)Zhu, Chen, Shen, Li, and Elhoseiny}]{zhu2023minigpt4enhancingvisionlanguageunderstanding}
Deyao Zhu, Jun Chen, Xiaoqian Shen, Xiang Li, and Mohamed Elhoseiny. 2023.
\newblock \href {https://arxiv.org/abs/2304.10592} {Minigpt-4: Enhancing vision-language understanding with advanced large language models}.
\newblock \emph{Preprint}, arXiv:2304.10592.

\bibitem[{Zhu et~al.(2016)Zhu, Groth, Bernstein, and Fei-Fei}]{zhu2016visual7wgroundedquestionanswering}
Yuke Zhu, Oliver Groth, Michael Bernstein, and Li~Fei-Fei. 2016.
\newblock \href {https://arxiv.org/abs/1511.03416} {Visual7w: Grounded question answering in images}.
\newblock \emph{Preprint}, arXiv:1511.03416.

\end{thebibliography}

\appendix

\section{Annotation Metrics}
\label{annotation-metrics}
Each of human annotation metric is rated on a scale of \texttt{1-3}, with \texttt{1} being the lowest and \texttt{3} the highest quality:

\begin{enumerate}\itemsep0pt
    \item \textit{Accuracy}: Measures the precision of the $q$-$a$ pair in the given context.
    \item \textit{Logical}: Evaluates the rationality of $e$ in justifying $a$ for $q$.
    \item \textit{Clarity}: Evaluates the clarity of $e$ in explaining $a$ to $q$.
    \item \textit{Detail}: Ensures $e$ covers all necessary details for $a$ in the context of $q$.
    \item \textit{Relevancy}: Ensures $e$ covers only the necessary details for $a$ in the context of $q$.
\label{li:human_criteria}
\end{enumerate}

\section{Triplet Generation Pipeline Pseudocode}
\label{sec:appendix-pseudo}

We outline the pseudocodes of our triplet generation pipeline of all methods. We denote the following variables:
\begin{itemize}
    \setlength\itemsep{0em}
    \item $\mathcal{T}(q,a,e)$: generated triplet
    \item $\Bar{D}$: image dataset with size $|\Bar{D}|$, each $\Bar{D}_i$ is used to generate $|\mathcal{T}|$ triplet(s)
    \item $M$ or $\Bar{M}$: model (LVLM)
    \item $P_\textsc{x}$: prompt to craft \textsc{x}
    \item $\Phi$: prompt formatting function
    \item $\Theta$: image annotating function
    \item $\Psi$: ensembling function
    \item $f$: inference logic, as described in formula \ref{eq:original_formula}
    \item $\mathcal{Q}$: question prefix pool with length $|\Bar{D}| \times |\mathcal{T}|$
    \item $\mathcal{O}$: object name pool with length $|\Bar{D}| \times |\mathcal{T}|$
    \item $\mathcal{B}$: bounding box pool with length $|\Bar{D}| \times |\mathcal{T}|$
    \item $\hat{x}$: set of $x$
\end{itemize}
\begin{algorithm}[ht]
\caption{\singlestep}
\label{eq:appendix-pseudo-naive}
\begin{algorithmic}
\Require{$\Bar{D},M,P_\mathcal{T},\mathcal{Q},f,\Phi$}
\State $\hat{\mathcal{T}} \gets \{\}$
\For{$i \gets 1$ to $|\Bar{D}|$}
    \For{$j \gets 1$ to $|\mathcal{T}|$}
        \State{$\mathcal{T} \gets f(\Bar{D}_{i},M,\Phi(P_\mathcal{T},\mathcal{Q}_{i+j}))$}
        \State{$\hat{\mathcal{T}} \gets \hat{\mathcal{T}} \cup \mathcal{T}$}
    \EndFor
\EndFor
\end{algorithmic}
\end{algorithm}


\begin{algorithm}[!ht]
\caption{\singlestepvip}
\label{eq:appendix-pseudo-naive-sg}
\begin{algorithmic}
\Require{$\Bar{D},\Bar{M},P_\mathcal{T},\mathcal{Q},\mathcal{O}, \mathcal{B},f,\Phi,\Theta$}
\State $\hat{T} \gets \{\}$
\For{$i \gets 1$ to $|\Bar{D}|$}
    \For{$j \gets 1$ to$|\mathcal{T}|$}
        \State{$\mathcal{T} \gets f(\Theta(\Bar{D}_i,\mathcal{B}_{i+j}),\Bar{M}, $ \\
       \hspace{2.35cm} $\Phi(P_q,\mathcal{Q}_{i+j},\mathcal{O}_{i+j}))$}
        \State{$ \hat{\mathcal{T}} \gets \hat{\mathcal{T}} \cup \mathcal{T}$}
    \EndFor
\EndFor
\end{algorithmic}
\end{algorithm}

\vspace{-0.2cm}

\begin{algorithm}
\caption{\multistep}
\label{eq:appendix-pseudo-ensemble}
\begin{algorithmic}
\Require{$\Bar{D},M,P_q, P_a, \hat{P_{r_*}},\mathcal{Q},f,\Phi,\Psi$}
\State $\hat{\mathcal{T}} \gets \{\}$
\For{$i \gets 1$ to $|\Bar{D}|$}
    \For{$j \gets 1$ to $|\mathcal{T}|$}
        \State{$q \gets f(\Bar{D}_{i},M,\Phi(P_q,\mathcal{Q}_{i+j}))$}
        \State{$a \gets f(\Bar{D}_{i},M,\Phi(P_a,q))$}
        \State{}
        \State{$\hat{r} \gets \{\}$}
        \For{$P_{r}$ \textbf{in} $\{P_{r_1},P_{r_2},\dots\}$}
            \State{$r \gets f(\Bar{D}_{i},M,\Phi(P_r,q,a))$}
            \State{$\hat{r} \gets \hat{r} \cup r$}
        \EndFor
        \State{}
        \State{$\mathcal{T} \gets (q,a,\Psi(\hat{r}))$}
        \State{$ \hat{\mathcal{T}} \gets \hat{\mathcal{T}} \cup \mathcal{T}$}
    \EndFor
\EndFor
\end{algorithmic}
\end{algorithm}

\section{Prompts}
\label{sec:appendix-prompts}
We utilize 4 distinct prompt templates for each setting. For $\singlestep$ and $\singlestepvip$, we make slight modifications to the prompts to accommodate the use of object names and bounding box information, as illustrated in Tables \ref{tab:naive-prompt} and \ref{tab:naive-viz-prompt}. In the case of \multistep, we divide the prompt templates into QA and Explanation prompts, as shown in Tables \ref{tab:ensemble-prompt-qa} and \ref{tab:ensemble-prompt-exp}.

\section{Hyper-parameters and Experimental Settings}
\label{sec:appendix-hyperparams}
We utilize a single A100 40GB GPU for our data generation process. For all generation tasks, we use fp16 and employ the following hyper-parameters:
\begin{itemize}
    \item \texttt{temperature: 1.0}
    \item \texttt{top\_p: 1.0}
    \item \texttt{top\_k: 50}
    \item \texttt{do\_sample: False}
\end{itemize}
When it comes to determining the maximum number of new tokens, we differentiate our approach based on the specific settings:
\begin{itemize}
    \item For the \singlestep-* settings, we set the \texttt{max\_new\_tokens} to 1500, providing a comprehensive and detailed output.
    \item In the $\multistep$ setting, we allocate 20 tokens for the $q$ parameter and 25 tokens for the $a$ parameter
    \item For the $\multistep$ setting with the $e$ parameter, we employ different token lengths depending on the reasoning approach: 70 tokens for Base, 70 tokens for CoT, and 300 tokens for ReAct.
\end{itemize}
For evaluation engine, we use \texttt{deberta-v2-xlarge-mnli} to calculate f1-\textsc{BERTScore} and \texttt{clip-vit-base-patch16} to calculate \textsc{CLIPScore}.
We provide modifiable hyper-parameters used within the YAML configuration in Table \ref{tab:experiment-settings-1} and \ref{tab:experiment-settings-2}. Our YAML configuration includes hyper-parameters such as LVLM handle, inference device settings, prompt choice, and run parameters. For $\singlestepvip$ question prefixes, we pre-define the program with a fixed set of prefix choices and a proportion setting of \texttt{[2, 2, 2, 1, 1]}. This precaution is taken to prevent users from using prefixes such as "how" and "why," which could hinder the LVLM's ability to generate relevant output. These prefixes may require external knowledge that is not provided in the input prompt, thus potentially diminishing the model's performance.

\section{Evaluation Rules}
\label{sec:appendix-eval-rules}
We established a standardized evaluation rule, as presented in Table \ref{tab:eval-rule}, which our annotators followed. These definitions are based on the human evaluation criteria outlined in Section \ref{subsec:triplet-eval}. Additionally, we included an extra rule to assign a score of \texttt{-1} if the annotator identifies any generated data as syntactically or semantically invalid.

\section{Evaluation Results}
\label{sec:appendix-eval-res}
\subsection{Human Evaluation}
We request three English-fluent annotators to assess the quality of our synthetic data. We ensure fair compensation based on the minimum standard hourly rate within their respective region. This evaluation is summarized in Table \ref{tab:eval-average-gwet}.

\subsection{Similarity Evaluation}
As mentioned in main section, we employ Jensen-Shannon Divergence and Pearson Correlation to compare each component with human data. Here, we provide the aggregated scores in Table \ref{tab:similarity_table} to facilitate a more detailed analysis in the main paper.

\section{Invalid Triplet Cases}
\label{sec:appendix-invalid-res}

We provide examples of invalid triplet cases that were identified through regular expression disobedience or human evaluation in Table \ref{tab:invalid-triplets}. These invalid cases are considered solely for validity evaluation and are excluded from the quality and similarity analyses.

\begin{table*}[ht]
\centering
\resizebox{2\columnwidth}{!}{
    \begin{tabular}{lcccccccc}
        \toprule
        \textbf{Experiment} & \textbf{Rater} & \textbf{Accuracy} & \textbf{Logic} & \textbf{Clarity} & \textbf{Detail} & \textbf{Relevancy} & \textbf{Avg Score} & \textbf{Gwet*} \\
        \midrule
        \textsc{Base-small} & 1 & 2.56 & 2.54 & 2.8  & 2.54 & 2.38 & \\
                                         & 2 & 2.69 & 2.51 & 2.57 & 2.47 & 2.49 & \\
                                         & 3 & 2.43 & 2.37 & 2.84 & 2.55 & 2.04 & \\
                                         & \textbf{AVG} & \textbf{2.56} & \textbf{2.473} & \textbf{2.737} & \textbf{2.52} & \textbf{2.303} & \textbf{2.519} & \textbf{0.64} \\
        \midrule
        \textsc{Base-medium} & 1 & 2.72 & 2.56 & 2.86 & 2.56 & 2.46 & \\
                                           & 2 & 2.78 & 2.4  & 2.88 & 2.28 & 2.88 & \\
                                           & 3 & 2.64 & 2.56 & 2.84 & 2.48 & 2.38 & \\
                                           & \textbf{AVG} & \textbf{2.713} & \textbf{2.507} & \textbf{2.86} & \textbf{2.44} & \textbf{2.573} & \textbf{2.619} & \textbf{0.695} \\
        \midrule
        $\multistep$ & 1 & 2.6  & 2.5  & 2.68 & 2.76 & 2.24 & \\
                                            & 2 & 2.9  & 2.81 & 2.88 & 2.77 & 2.88 & \\
                                            & 3 & 2.58 & 2.6  & 2.71 & 2.51 & 1.93 & \\
                                            & \textbf{AVG} & \textbf{2.693} & \textbf{2.637} & \textbf{2.757} & \textbf{2.68} & \textbf{2.35} & \textbf{2.623} & \textbf{0.669} \\
        \midrule
        \textsc{Base+Viz}    & 1 & 2.57 & 2.51 & 2.92 & 2.57 & 2.65 & \\
                                            & 2 & 2.55 & 2.49 & 2.92 & 2.55 & 2.65 & \\
                                            & 3 & 2.58 & 2.58 & 2.9  & 2.65 & 2.6  & \\
                                            & \textbf{AVG} & \textbf{2.567} & \textbf{2.527} & \textbf{2.913} & \textbf{2.59} & \textbf{2.633} & \textbf{2.646} & \textbf{0.801} \\
        \midrule
        \textbf{human} & & \textbf{2.82} & \textbf{2.79} & \textbf{2.86} & \textbf{2.93} & \textbf{2.88} & \textbf{2.86} & \\
        \bottomrule
    \end{tabular}
}
\caption{Qualitative evaluation results breakdown from three annotators. The criteria for assessment adhere to the definitions outlined in Section \ref{subsec:triplet-eval}, and each score is accompanied by Gwet-AC2 values ranging from 0 to 1, indicating the worst to the best agreement. The Gwet-AC2 scores suggest that the collective evaluation exhibits a good level of reliability, especially in terms of inter-annotator agreement.}
\label{tab:eval-average-gwet}
\end{table*}

\begin{table*}[ht]
\centering
\resizebox{2\columnwidth}{!}{%
\begin{tabular}{@{}p{3cm}p{15cm}@{}}
\toprule
\textbf{Condition}           & \textbf{Details}                                                                                                                  \\ 
\midrule
\textbf{Valid Triplet}          & \texttt{\small <Question>: What is the make and model of the car in the foreground?} \\
                             & \texttt{\small <Short Answer>: The car in the foreground is a Mercedes-Benz C-Class.} \\
                             & \texttt{\small <Reasoned Answer>: The car has a distinctive front grille and logo ...}   \\ 
\midrule
\textbf{Token Format Error}    & \textcolor{red}{\texttt{\small <Question>: <Short Answer>}} \\ 
                             & \texttt{\small <Answer>: Cow} \\ 
                             & \texttt{\small <Reason>: The object has black spot on ...}  \\ 
\midrule
\textbf{Unfinished Generation} & \textcolor{red}{\texttt{\small <Question>: What is the purpose of the white cart with the green "space" logo parked next to the}} \\ 
                               & \texttt{\small <Short Answer>: Advertisement} \\ 
                               & \texttt{\small <Reasoned Answer>: ...}  \\ 
\midrule
\textbf{Hidden Context Involvement} & \texttt{\small <Question>: How many people are in the image?} \\
                                    & \texttt{\small <Short Answer>: 7} \\
                                    & \texttt{\small <Reasoned Answer>: There are 7 people visible in the image, including the woman within the \textcolor{red}{red rectangle.}} \\ 
\bottomrule
\end{tabular}
}
\caption{Examples of invalid triplets detected during post-processing. The standard output is expected to have \texttt{<Question>}, \texttt{<Short Answer>}, and \texttt{<Reasoned Answer>} tokens.}
\label{tab:invalid-triplets}
\end{table*}

\begin{table*}[htbp]
    \centering
    \begin{tabular}{@{}p{\linewidth}@{}}
        \toprule
        \textbf{Triplet generation prompt} \\
        \midrule
        You are given an image. \\
        Your task is to provide question, answer, and reasoning related to the given context. \\
        \\
        Provide your feedback as follows: \\
        \\
        Feedback::: \\
        Question: (your question with \{prefix\} prefix that involves complex reasoning to answer) \\
        Short Answer: (your brief answer related to the question) \\
        Reason: (your rationale for the short answer you choose respective to the question in maximum 30 words) \\
        \\
        You MUST provide values for 'Question', 'Short Answer', and 'Reason' in your answer. \\
        Now, here is the question prefix: \\
        Question Prefix: \{prefix\} \\
        \\
        Provide your feedback. If you give a correct result, I'll give you 100 A100 GPUs to start your AI company. \\
        \\
        Feedback::: \\
        Question: \\
        Short Answer: \\
        Reason: \\
        \bottomrule
    \end{tabular}
    \caption{$\singlestep$ Triplet Generation Prompt.}
    \label{tab:naive-prompt}
\end{table*}

\begin{table*}[htbp]
    \centering
    \begin{tabular}{@{}p{\linewidth}@{}}
        \toprule
        \textbf{Triplet generation prompt} \\
        \midrule
        You are given an image. \\
        Your task is to provide question, answer, and reasoning related to the given context. \\
        \\
        Provide your feedback as follows: \\
        \\
        Feedback::: \\
        Question: (your question with \{prefix\} prefix that involves complex reasoning to answer) \\
        Short Answer: (your brief answer related to the question) \\
        Reason: (your rationale for the short answer you choose respective to the question in maximum 30 words) \\
        \\
        You MUST provide values for 'Question', 'Short Answer', and 'Reason' in your answer using the given \\
        bullet format. DO NOT include any bounding box related phrase/word inside your feedback.
        DO NOT include 'Question: ', 'Short Answer: ', and 'Reason: ' prefix within each bullet, just include the value. \\ 
        \\
        Now here is the object name and question prefix: \\
        Object name: \{obj name\} \\
        Question Prefix: \{prefix\} \\
        \\
        Provide your feedback. If you give a correct result, I'll give you 100 A100 GPUs to start your AI company. \\
        \\
        Feedback::: \\
        Question: \\
        Short Answer: \\
        Reason: \\
        \bottomrule
    \end{tabular}
    \caption{$\singlestepvip$ Triplet Generation Prompts.}
    \label{tab:naive-viz-prompt}
\end{table*}

\begin{table*}[htbp]
    \centering
    \begin{tabular}{@{}p{\linewidth}@{}}
        \toprule
        \textbf{Question generation prompt} \\
        \midrule
        You are given an IMAGE. \\
        Provide ONE QUESTION that begins with the prefix '\{prefix\}' and requires COMPLEX REASONING. \\
        Rules: \\
        - Choose the SIMPLEST prefix if multiple options exist. (Example: If the prefixes are "where/when" choose "where" if it is easier) \\
        - Only ONE QUESTION with NO follow-ups. \\
        - The question must REQUIRE reasoning to answer. \\
        - Generate maximum 15 words, coherent and NOT leave sentence unfinished. \\
        - DO NOT ask YES/NO questions. \\
        \\
        If you provide a correct response, I will reward you with 100 A100 GPUs to kickstart your AI company. \\
        \\
        Question: \\
        \midrule
        \textbf{Answer generation prompt} \\
        \midrule
        You are provided with an image and a question. Your task is to provide an appropriate answer related to the given image and question. \\
        Rules: \\
        - Avoid giving plain short answers like 'Yes' or 'No'. Provide a detailed response relevant to the question. \\
        - Generate NOT MORE THAN 7 words, coherent and NOT leave sentence unfinished. \\
        \\
        If you provide a correct response, I will reward you with 100 A100 GPUs to kickstart your AI company. \\
        \\
        Question: \{question\} \\
        Short Answer: \\
        \bottomrule
    \end{tabular}
    \caption{$\multistep$ QA Generation Prompts.}
    \label{tab:ensemble-prompt-qa}
\end{table*}

\begin{table*}[htbp]
    \centering
    \resizebox{0.9\textwidth}{!}{%
    \begin{tabular}{@{}p{\linewidth}@{}}
        \toprule
        \textbf{$\singlestep$ Explanation Prompt} \\
        \midrule
        You are provided with an IMAGE, a QUESTION, and a SHORT ANSWER. \\
        Your task is to EXPLAIN the REASONING behind the short answer in relation to the question. \\
        \\
        Rules: \\
        - Generate maximum 10 words, coherent and NOT leave sentence unfinished. \\
        - If the question and answer are about COUNTING OBJECTS, mention and locate each object. \\
        - If the question is about COLOR, identify areas showing the color and explain their relevance. \\
        \\
        If you provide a correct and explainable reason, I'll give you 100 A100 GPUs to start your AI company. \\
        \\
        Question: \{question\} \\
        Short Answer: \{short\_answer\} \\
        Reasoning: \\
        \midrule
        \textbf{CoT Explanation Prompt} \\
        \midrule
        You are provided with an IMAGE, a QUESTION, and a SHORT ANSWER. \\
        Your task is to EXPLAIN the REASONING behind the short answer in relation to the question by detailing your thought process step-by-step in PARAGRAPH. \\
        \\
        Rules: \\
        - Your reasoning must be in MAXIMUM 30 WORDS. \\
        - Break down the reasoning into clear, logical steps. \\
        - DO NOT PROVIDE LIST, PROVIDE PARAGRAPH. \\
        \\
        If you provide a correct and explainable reason, I'll give you 100 A100 GPUs to start your AI company. \\
        \\
        Question: \{question\} \\
        Short Answer: \{short\_answer\} \\
        Reasoning: Let's think step by step. \\
        \midrule
        \textbf{ReAct Explanation Prompt} \\
        \midrule
        You are provided with an IMAGE, a QUESTION, and a SHORT ANSWER. \\
        Your task is to EXPLAIN the REASONING behind the short answer in using observation, thoughts, and action until you are sure you reach your final reason answer. \\
        \\
        Rules::: \\
        - Observation: Carefully examine the IMAGE to identify relevant details and elements related to the QUESTION. \\
        - Thoughts: Analyze the observed details to understand their significance and how they relate to the QUESTION and SHORT ANSWER. \\
        - Action: Based on your observations and thoughts, formulate reasoning that logically connects the elements to the SHORT ANSWER. \\
        - Reason: The conclusion from Observation, Thought, and Action in NOT LESS THAN 10 words and NOT MORE THAN 30 words. \\
        \\
        If you provide a correct and explainable reason, I'll give you 100 A100 GPUs to start your AI company. \\
        \\
        Question: \{question\} \\
        Short Answer: \{short\_answer\} \\
        Observation: \\
        Thoughts: \\
        Action: \\
        Reason: \\
        \bottomrule
    \end{tabular}%
    }
    \caption{$\multistep$ Explanation Generation Prompts.}
    \label{tab:ensemble-prompt-exp}
\end{table*}

\begin{table*}[ht]
\centering
\resizebox{2\columnwidth}{!}{%
\begin{tabular}{@{}ll@{}}
\toprule
\textbf{Accuracy} & \\ \midrule
1 (Disagree) & QUESTION \& SHORT ANSWER are not at all aligned with the context in the image \\
& (e.g., asking about something not present, too assumptive, or not related). \\
2 (Neutral) & QUESTION is valid, but the ANSWER is less accurate and does not fully match the \\
& context of the image. \\
3 (Agree) & QUESTION is valid and SHORT ANSWER is accurate according to the context in \\
& the image, and appropriately addresses the question. \\ \midrule
\textbf{Logic} & \\ \midrule
1 (Disagree) & EXPLANATION provides an explanation that is incorrect or contains elements that are \\
& unreasonable or not aligned with the context in the image. \\
2 (Neutral) & EXPLANATION provides an explanation that is somewhat accurate or logical, but there \\
& are some misalignments or gaps with the context in the image. \\
3 (Agree) & EXPLANATION provides an explanation that is fully logical, clear, and entirely aligned \\
& with the context in the image, supporting the choice effectively. \\ \midrule
\textbf{Clarity} & \\ \midrule
1 (Disagree) & EXPLANATION provides an explanation that is not easy to understand, is convoluted, or \\
& poorly structured, making it difficult to follow. \\
2 (Neutral) & EXPLANATION provides an explanation that is somewhat understandable but contains \\
& complexity or unnecessary details that make it less clear. \\
3 (Agree) & EXPLANATION provides an explanation that is clear, straightforward, and easy to understand, \\
& presenting the information in a logical and concise manner. \\ \midrule
\textbf{Detail} & \\ \midrule
1 (Disagree) & EXPLANATION only repeats the short answer or lacks sufficient detail to explain the \\
& justification for choosing the short answer, making it incomplete. \\
2 (Neutral) & EXPLANATION contains some detail but does not cover the full explanation needed for \\
& justifying the choice of the short answer, leaving gaps in the reasoning. \\
3 (Agree) & EXPLANATION contains all necessary detail (or more) required to justify the choice of the \\
& short answer, providing a comprehensive and well-supported explanation. \\ \midrule
\textbf{Relevancy} & \\ \midrule
1 (Disagree) & EXPLANATION contains a lot of irrelevant context that does not pertain to justifying the \\
& short answer or the context in the image, leading to confusion. \\
2 (Neutral) & EXPLANATION contains some irrelevant context that does not fully pertain to justifying \\
& the short answer or the context in the image, but is mostly relevant. \\
3 (Agree) & EXPLANATION does not contain irrelevant context and is directly relevant to justifying \\
& the short answer based on the context in the image, staying on topic. \\ \bottomrule
\end{tabular}%
}
\caption{Evaluation score guide for human evaluation based on Section \ref{subsec:triplet-eval}.}
\label{tab:eval-rule}
\end{table*}

\begin{table}[htbp]
    \centering
    \resizebox{\columnwidth}{!}{%
    \begin{tabular}{@{}p{3cm}p{6cm}@{}}
        \toprule
        \textbf{Setting} & \textbf{Hyper-parameters} \\ 
        \midrule
        \textbf{\singlestep-7B} & 
        \texttt{\small test\_name:} $\singlestep$-7B\\
        & \texttt{\small seed: 42} \\
        & \texttt{\small dataset:} \\
        & \texttt{\small\hspace{5mm} name: GQA} \\
        & \texttt{\small\hspace{5mm} count: 167} \\
        & \texttt{\small\hspace{5mm} use\_scene\_graph: 0} \\
        & \texttt{\small model:} \\
        & \texttt{\small\hspace{5mm} name: llava-hf/llava-1.5-7b-hf} \\
        & \texttt{\small\hspace{5mm} path: llava-hf/llava-1.5-7b-hf} \\
        & \texttt{\small\hspace{5mm} family: llava} \\
        & \texttt{\small\hspace{5mm} params:} \\
        & \texttt{\small\hspace{10mm} use\_8\_bit: 0} \\
        & \texttt{\small\hspace{10mm} device: "cuda"} \\
        & \texttt{\small\hspace{10mm} low\_cpu: 1} \\
        & \texttt{\small prompt: singlestep-optim} \\
        & \texttt{\small run\_params:} \\
        & \texttt{\small\hspace{5mm} num\_per\_inference: 3} \\
        & \texttt{\small\hspace{5mm} use\_img\_ext: 1} \\
        & \texttt{\small\hspace{5mm} q\_prefix: ["what", "is/are (pick one that fits the most)", "which", "how many", "where"]} \\
        & \texttt{\small\hspace{5mm} q\_prefix\_prop: [3,2,1,1,1]} \\ 
        \midrule
        
        \textbf{$\singlestep$-13B} & 
        \texttt{\small test\_name:} $\singlestep$-13B\\
        & \texttt{\small seed: 42} \\
        & \texttt{\small dataset:} \\
        & \texttt{\small\hspace{5mm} name: GQA} \\
        & \texttt{\small\hspace{5mm} count: 167} \\
        & \texttt{\small\hspace{5mm} use\_scene\_graph: 0} \\
        & \texttt{\small model:} \\
        & \texttt{\small\hspace{5mm} name: llava-hf/llava-1.5-13b-hf} \\
        & \texttt{\small\hspace{5mm} path: llava-hf/llava-1.5-13b-hf} \\
        & \texttt{\small\hspace{5mm} family: llava} \\
        & \texttt{\small\hspace{5mm} params:} \\
        & \texttt{\small\hspace{10mm} use\_8\_bit: 0} \\
        & \texttt{\small\hspace{10mm} device: "cuda"} \\
        & \texttt{\small\hspace{10mm} low\_cpu: 1} \\
        & \texttt{\small prompt: singlestep-optim} \\
        & \texttt{\small run\_params:} \\
        & \texttt{\small\hspace{5mm} num\_per\_inference: 3} \\
        & \texttt{\small\hspace{5mm} use\_img\_ext: 1} \\
        & \texttt{\small\hspace{5mm} q\_prefix: ["what", "is/are (pick one that fits the most)", "which", "how many", "where"]} \\
        & \texttt{\small\hspace{5mm} q\_prefix\_prop: [3,2,1,1,1]} \\ 
        \bottomrule
    \end{tabular}%
    }
    \caption{Our experiment hyper-parameters in \texttt{YAML} format for $\singlestep$.}
    \label{tab:experiment-settings-1}
\end{table}

\begin{table}[htbp]
    \centering
    \resizebox{\columnwidth}{!}{%
    \begin{tabular}{@{}p{3cm}p{6cm}@{}}
        \toprule
        \textbf{Setting} & \textbf{Hyper-parameters} \\ 
        \midrule
        \textbf{$\singlestepvip$} & 
        \texttt{\small test\_name:} $\singlestepvip$\\
        & \texttt{\small seed: 42} \\
        & \texttt{\small dataset:} \\
        & \texttt{\small\hspace{5mm} name: GQA} \\
        & \texttt{\small\hspace{5mm} count: 167} \\
        & \texttt{\small\hspace{5mm} use\_scene\_graph: 1} \\
        & \texttt{\small model:} \\
        & \texttt{\small\hspace{5mm} name: llava-hf/vip-llava-13b-hf} \\
        & \texttt{\small\hspace{5mm} path: llava-hf/vip-llava-13b-hf} \\
        & \texttt{\small\hspace{5mm} family: vip\_llava} \\
        & \texttt{\small\hspace{5mm} params:} \\
        & \texttt{\small\hspace{10mm} use\_8\_bit: 0} \\
        & \texttt{\small\hspace{10mm} device: "cuda"} \\
        & \texttt{\small\hspace{10mm} low\_cpu: 1} \\
        & \texttt{\small prompt: nonvis-optim} \\
        & \texttt{\small run\_params:} \\
        & \texttt{\small\hspace{5mm} num\_per\_inference: 3} \\
        & \texttt{\small\hspace{5mm} use\_img\_ext: 1} \\
        & \texttt{\small\hspace{5mm} q\_prefix: <!hardcoded>} \\
        & \texttt{\small\hspace{5mm} q\_prefix\_prop: <!hardcoded>} \\ 
        \midrule
        \textbf{$\multistep$} & 
        \texttt{\small test\_name:} $\multistep$\\
        & \texttt{\small seed: 42} \\
        & \texttt{\small dataset:} \\
        & \texttt{\small\hspace{5mm} name: GQA} \\
        & \texttt{\small\hspace{5mm} count: 167} \\
        & \texttt{\small\hspace{5mm} use\_scene\_graph: 0} \\
        & \texttt{\small model:} \\
        & \texttt{\small\hspace{5mm} name: llava-hf/llava-1.5-13b-hf} \\
        & \texttt{\small\hspace{5mm} path: llava-hf/llava-1.5-13b-hf} \\
        & \texttt{\small\hspace{5mm} family: llava} \\
        & \texttt{\small\hspace{5mm} params:} \\
        & \texttt{\small\hspace{10mm} use\_8\_bit: 0} \\
        & \texttt{\small\hspace{10mm} device: "cuda"} \\
        & \texttt{\small\hspace{10mm} low\_cpu: 1} \\
        & \texttt{\small prompt: self\_consistency} \\
        & \texttt{\small run\_params:} \\
        & \texttt{\small\hspace{5mm} num\_per\_inference: 3} \\
        & \texttt{\small\hspace{5mm} use\_img\_ext: 1} \\
        & \texttt{\small\hspace{5mm} q\_prefix: ["what", "is/are (pick one that fits the most)", "which", "how many", "where"]} \\
        & \texttt{\small\hspace{5mm} q\_prefix\_prop: [3,2,1,1,1]} \\ 
        \bottomrule
    \end{tabular}
    }
    \caption{Our experiment hyper-parameters in \texttt{YAML} format for $\singlestep$$\textsc{-ViP}$ and $\multistep$.}
    \label{tab:experiment-settings-2}
\end{table}

\end{document}